\documentclass[hidelinks,nohypdvips,onefignum,onetabnum]{siamart250211}

\usepackage[T1]{fontenc}
\usepackage[utf8]{inputenc}
\usepackage{lmodern}
\ifdefined\pdfgentounicode
\input{glyphtounicode.tex}
\fi
\usepackage{amsfonts,amssymb,mathtools,bm}
\usepackage{graphicx,booktabs,array,multirow,placeins}
\usepackage{algorithm,algorithmic}

\makeatletter
\renewcommand\section{\@startsection{section}{1}{0pt}%
                                   {2.0ex \@plus .5ex \@minus .2ex}%
                                   {0.8ex \@plus .2ex}%
                                   {\reset@font\large\bfseries}}
\renewcommand\subsection{\@startsection{subsection}{2}{0pt}%
                                     {1.6ex \@plus .4ex \@minus .2ex}%
                                     {0.6ex \@plus .2ex}%
                                     {\reset@font\normalsize\bfseries}}
\renewcommand\subsubsection{\@startsection{subsubsection}{3}{0pt}%
                                        {1.3ex \@plus .3ex \@minus .2ex}%
                                        {0.4ex \@plus .1ex}%
                                        {\reset@font\normalsize\bfseries}}
\renewcommand\paragraph{\@startsection{paragraph}{4}{0pt}%
                                     {1.0ex \@plus .2ex \@minus .1ex}%
                                     {-0.5em \@plus -.1em}%
                                     {\reset@font\normalsize\bfseries}}
\makeatother

\newsiamremark{remark}{Remark}
\newsiamremark{note}{Note}
\crefname{remark}{Remark}{Remarks}
\crefname{note}{Note}{Notes}

\newcommand{\ofs}[1]{\mathbf{T}^{\text{ofs}}_{#1}}
\newcommand{\ifi}[1]{\mathbf{T}^{\text{ifi}}_{#1}}
\newcommand{\ofo}[1]{\mathbf{T}^{\text{ofo}}_{#1}}
\newcommand{\tfi}[1]{\mathbf{T}^{\text{tfi}}_{#1}}
\newcommand{\ifo}[1]{\mathbf{T}^{\text{ifo}}_{#1}}
\newcommand{\Tau}{\mathcal{T}}

\headers{Learning Green's Operators via GridNFMM}{E. McAllister Fognini, M. M. Betcke, and B. T. Cox}

\title{Learning Green's Operators through Hierarchical Neural Networks Inspired by the Fast Multipole Method%
\thanks{This manuscript revises and expands the preprint \emph{Learning Green's Operators through Hierarchical Neural Networks Inspired by the Fast Multipole Method} (arXiv:2509.20591v1).}}

\author{Emilio McAllister Fognini\thanks{Department of Computer Science, University College London, London, United Kingdom (\email{emilio.fognini.17@ucl.ac.uk}, \email{m.betcke@ucl.ac.uk}).}
\and Marta M.~Betcke\footnotemark[2]
\and Ben T.~Cox\thanks{Department of Medical Physics and Biomedical Engineering, University College London, London, United Kingdom (\email{b.cox@ucl.ac.uk}).}}

\hypersetup{
  pdftitle={Learning Green's Operators through Hierarchical Neural Networks Inspired by the Fast Multipole Method},
  pdfauthor={Emilio McAllister Fognini, Marta M. Betcke, and Ben T. Cox},
  pdfsubject={Machine Learning Methods for Scientific Computing},
  pdfkeywords={Fast Multipole Method, Hierarchical Neural Networks, Operator Learning, Helmholtz Equation, Elliptic PDEs}
}

\begin{document}

\maketitle

\begin{abstract}
  The Fast Multipole Method (FMM) is an efficient numerical algorithm for evaluating long-range interactions in $N$-body problems and integral equations in $\mathcal{O}(N)$ operations by combining hierarchical domain decomposition with low-rank multipole expansions of the underlying Green's function.
  Despite its widespread application in computational physics and engineering, classical FMM formulations require an explicitly known, analytically tractable interaction kernel, limiting their direct application to problems in heterogeneous media where no closed-form Green's function exists.
  In this work, we follow the design principle \emph{fix the computational graph, learn the computation} to propose the \textbf{Grid-based Neural Fast Multipole Method (GridNFMM)}, a hierarchical neural architecture that embeds the multi-level information flow and near/far-field division of labour of the FMM into a deep operator-learning framework.
  By replacing kernel-specific multipole and local translation operators with learnable factorised message-passing layers across the FMM interaction lists $\mathcal{U}_\tau$, triangular inter-level shifts, and a factorised $3 \times 3$ near-field convolutional operator, the GridNFMM learns solution operators directly from data while retaining $\mathcal{O}(N)$ computational complexity per block.
  We evaluate the GridNFMM against the Tucker-Factorised Fourier Neural Operator (TFNO) across standard 2D PDE benchmarks (Darcy Flow and Vorticity Navier--Stokes) and three custom 2D heterogeneous Helmholtz scattering regimes.
  Our results demonstrate that the GridNFMM outperforms the TFNO across all evaluated Helmholtz scattering regimes and Darcy Flow, mitigating the spectral bias of truncated Fourier representations on oscillatory wavefields.
\end{abstract}



% =============================================================================
\section{Introduction}\label{sec:introduction}
% =============================================================================

Simulations of time-harmonic acoustic wave scattering in media with spatially varying wave speed underpin numerous scientific and engineering applications, including biomedical ultrasound tomography, seismic imaging, non-destructive testing, and sonar.
In these settings, computational efficiency governs not only the cost of a single forward simulation, but also the feasibility of solving inverse problems or exploring material configurations that require repeated evaluations of partial differential equations (PDEs) \cite{colton_inverse_2019}.
Oscillatory and high-frequency wave problems governed by the Helmholtz equation are particularly demanding: resolving short wavelengths requires fine spatial discretisations, while the resulting linear systems are generally non-coercive and indefinite, hindering iterative solvers and necessitating additional mesh refinement to control numerical dispersion and the pollution effect \cite{babuska_is_1997, christodoulou_high-order_2017}.

Classical numerical analysis addresses the cost of dense or large-scale linear systems through fast algorithms that exploit structural properties of the underlying operators, such as locality, sparsity, separability, and off-diagonal low-rank structure \cite{martinsson_fast_2019}.
Prominent among these is the Fast Multipole Method (FMM) \cite{rokhlin_rapid_1985, greengard_fast_1987}, which reduces the complexity of evaluating pairwise $N$-body interactions from $\mathcal{O}(N^2)$ to $\mathcal{O}(N)$ (or $\mathcal{O}(N \log N)$ for high-frequency oscillatory kernels \cite{coifman_fast_1993, cheng_wideband_2006, engquist_fast_2007}) by partitioning the domain hierarchically: near-field interactions are evaluated directly, whereas far-field interactions between well-separated regions are compressed and translated through hierarchical tree traversals.
However, conventional FMM algorithms require an analytically available interaction kernel, or a free-space fundamental solution in volume integral formulations coupled with iterative Krylov solvers \cite{ying_kernel-independent_2004, ying_pedestrian_2012}.
For general heterogeneous media or bounded domains with complex boundary conditions, closed-form Green's functions are unavailable, and changing the medium properties or wavenumber typically requires re-solving or re-factorising the discretised system.

In parallel, operator learning and Neural Operators (NOs) \cite{lu_learning_2021, kovachki_neural_2021, kovachki_operator_2024} have emerged as a data-driven framework for learning mappings between infinite-dimensional function spaces directly from observations.
Yet standard neural operators face complementary challenges on oscillatory PDEs.
Deep neural networks exhibit a well-documented \emph{spectral bias} toward low-frequency functions \cite{rahaman_spectral_2019}, and prominent spectral architectures such as the Fourier Neural Operator (FNO) \cite{li_fourier_2021} and its tensorised variants (TFNO) \cite{kossaifi_multi-grid_2023} parametrise global integral kernels via a truncated Fourier basis, acting as a low-pass filter in their spectral branch that can oversmooth high-frequency wavefields and sharp material interfaces.

In this paper, we bridge classical fast summation algorithms and modern operator learning through the design principle: \emph{fix the computational graph, learn the computation}.
Rather than asking a neural network to discover both an efficient interaction topology and the operator approximation simultaneously, we prescribe the hierarchical domain decomposition and multi-level information flow of the FMM while learning the constituent local and far-field operators from data.
Specifically, we introduce the \textbf{Grid-based Neural Fast Multipole Method (GridNFMM)}\footnote{Referred to simply as the Neural FMM in an earlier version of this preprint; we adopt the name GridNFMM here to distinguish this fixed-grid architecture from continuous point-cloud extensions.}, a tree-based neural architecture that replaces handcrafted multipole and local translation operators with learnable neural modules operating on a Morton-ordered quadtree.

\paragraph{Our contribution}
We extend the algorithmic structure of the FMM to scientific machine learning by leveraging its non-local hierarchical information flow and near/far-field spatial partitioning within a deep operator-learning architecture.
We formulate the GridNFMM block, which structurally encodes the upward pass, downward pass, and near/far-field division of labour of the classical FMM on a uniform quadtree.
Unlike multi-scale U-Net or V-cycle graph architectures that rely on overlapping receptive fields, the GridNFMM explicitly isolates the non-overlapping FMM interaction lists $\mathcal{U}_\tau$ at each tree level and pairs them with a factorised $3 \times 3$ near-field convolutional operator at the leaf level.
Exploiting the translation invariance of the uniform quadtree interaction list, which admits at most $K = 40$ unique relative displacement vectors per level in two dimensions, we introduce a parameter-efficient factorised parametrisation for the Multipole-to-Local ($\ifo{\tau,\sigma}$) operator combined with zero-initialised multiplicative gating, and prove that a forward pass of the GridNFMM block requires $\mathcal{O}(N c^2)$ floating-point operations and $\mathcal{O}(N c)$ memory—scaling linearly as $\mathcal{O}(N)$ in the number of grid points $N$ for fixed channel width $c$.
Through systematic benchmarks against a parameter-matched Tucker-Factorised FNO (TFNO) on 2D Darcy Flow, Vorticity Navier--Stokes, and three custom 2D heterogeneous Helmholtz scattering regimes spanning weak scattering, transmission, and hard scattering across fixed and variable media and sources, we demonstrate that the GridNFMM consistently outperforms the TFNO on elliptic and oscillatory wave scattering problems.
An open-source PyTorch implementation of the GridNFMM is available at \url{https://github.com/Emilio4d46/NeuralFMM}.

% =============================================================================
\section{Related work and preliminaries}\label{sec:preliminaries}
% =============================================================================

\subsection{Neural Operators}
Operator learning seeks to approximate mappings $\mathcal{G}^\dagger : \mathcal{A} \to \mathcal{U}$ between infinite-dimensional function spaces, such as the parameter-to-solution or source-to-solution maps of PDEs \cite{lu_learning_2021, kovachki_neural_2021, kovachki_operator_2024, berner_principled_2025}.
Throughout this paper, $\mathcal{D} \subset \mathbb{R}^d$ ($d \in \{2, 3\}$) denotes a bounded spatial domain with coordinates $r, r' \in \mathcal{D}$.
When a field is evaluated at $N$ discrete points $\{r_i\}_{i=1}^N \subset \mathcal{D}$, each point $i$ carries a column feature vector $x_i \in \mathbb{R}^c$ of channel dimension $c$, stacked row-wise into the data matrix $\mathbf{X} = [x_1, \dots, x_N]^\top \in \mathbb{R}^{N \times c}$.
We distinguish between \emph{pointwise channel mixers}, which act independently at each spatial location $i$ across its $c$ channels, \emph{spatial mixers}, which couple information across distinct spatial locations $j \neq i$, and \emph{latent representations}, which denote spatially compressed states associated with coarse boxes in a hierarchical tree.
For pointwise transformations, we denote a strictly linear (bias-free) weight matrix $\mathbf{W}_* \in \mathbb{R}^{c_{\text{out}} \times c_{\text{in}}}$ acting on a column vector as $\mathbf{W}_* x_i$ (or $\mathbf{X}\mathbf{W}_*^\top$ row-wise), and an affine transformation carrying a learnable bias vector $b_* \in \mathbb{R}^{c_{\text{out}}}$ with a superscript bullet as $\mathbf{W}_*^\bullet(x_i) := \mathbf{W}_* x_i + b_*$, where each weight matrix and bias is scoped locally to its enclosing operator.

Building on the universal operator approximation framework of DeepONet \cite{chen_universal_1995, lu_learning_2021}, Kovachki et al.~\cite{kovachki_neural_2021} formalised Neural Operators composed of a pointwise lifting operator $\mathbf{P}$, $T$ operator blocks $\{\mathcal{B}_t\}_{t=1}^T$, and a pointwise projection operator $\mathbf{Q}$:
\begin{equation}\label{eqn:Neural_Operator_Unrolled}
  \mathcal{G}_\theta(a)(r) = \left( \mathbf{Q} \circ \mathcal{B}_T \circ \dots \circ \mathcal{B}_1 \circ \mathbf{P}(a) \right)(r).
\end{equation}
In the classical formulation, each block $\mathcal{B}_t$ updates the hidden representation $v_{t-1} : \mathcal{D} \to \mathbb{R}^{c_v}$ via a pointwise linear transformation $\mathbf{W}_t$, a non-local integral spatial mixer $\mathcal{K}_t$, and an elementwise non-linearity $\sigma_t$:
\begin{equation}\label{eqn:Neural_Operator_Step}
  v_t(r) = \sigma_t\left(\mathbf{W}_{t} v_{t-1}(r) + \left(\mathcal{K}_{t}(v_{t-1}; a)\right)(r) + b_{t}\right),
\end{equation}
where the integral kernel operator takes the general form
\begin{equation}\label{eqn:Neural_Operator_Non-Local_Operator}
  \left(\mathcal{K}_t(v_{t-1}; a)\right)(r) = \int_{\mathcal{D}} \kappa^{(t)}\!\left(r, r', a(r), a(r'), v_{t-1}(r), v_{t-1}(r')\right) v_{t-1}(r')\, d\nu(r').
\end{equation}
Modern transformer-style neural operators frequently decouple the spatial mixer $\mathcal{K}_t$ from a subsequent pointwise channel-mixing network ($\operatorname{DNN}_t$, such as a Gated Linear Unit, GLU \cite{dauphin2017language, shazeer_glu_2020}) using pre-LayerNorm \cite{ba_layer_2016} residual sub-blocks:
\begin{align}\label{eqn:New_Neural_Operator_Step}
  h(r) &= v_{t-1}(r) + \left(\mathcal{K}_t\big(\mathrm{LayerNorm}(v_{t-1}); a\big)\right)(r), \nonumber \\
  v_t(r) &= h(r) + \operatorname{DNN}_t\big(\mathrm{LayerNorm}(h(r))\big).
\end{align}
As classified by Boullé and Townsend \cite{boulle_mathematical_2023}, structural assumptions on the kernel $\kappa^{(t)}$ correspond to structured matrix classes upon discretisation: low-rank (such as Low-Rank Neural Operators, Galerkin and Fourier Transformers \cite{cao_choose_2021, li_transformer_2023}, Transolver \cite{wu_transolver_2024, luo_transolver_2025}, and LinearNO \cite{hu_transolver_2026}), local and sparse (such as Graph Neural Operators \cite{li_neural_2020} and DISCO convolutions \cite{ocampo_scalable_2023}), spectral and circulant, and hierarchical off-diagonal low-rank.
We focus primarily on the Fourier Neural Operator (FNO) family as the fixed-grid benchmark for the GridNFMM.

\subsection{Related work}\label{sec:related_work}

\paragraph{Inductive priors and hierarchical matrix neural networks}
For oscillatory wave scattering governed by the Helmholtz equation, strong inductive priors have been explored to aid operator learning.
Stanziola et al.~\cite{stanziola_learned_2023} parameterise a learned convergent Born series for heterogeneous media, while Fan et al.~\cite{fan_bcr-net_2019} introduced BCR-Net based on the non-standard wavelet form for linear integral operators, applying it to inverse Helmholtz scattering \cite{fan_solving_2019}.
Closely related are multiscale neural networks that adapt hierarchical matrices ($\mathcal{H}$-matrices \cite{fan_multiscale_2019}) and nested $\mathcal{H}^2$-bases (MNN-$\mathcal{H}^2$ \cite{fan_nested_multiscale_2019}) into deep learning architectures \cite{Hackbusch1999, martinsson_fast_2019}.
Within the graph and transformer literature, the Multipole Graph Neural Operator (MGNO) \cite{li_multipole_2020} employs a multi-level V-cycle over subsampled radius graphs, while the Multigrid Neural Operator (MgNO) \cite{he_mgno_2024}, Hierarchical Attention Neural Operators (HANO) \cite{liu_mitigating_2024, luo_hierarchical_2024}, and FMMformer \cite{nguyen_fmmformer_2021} decompose interactions across scales.
However, architectures such as MGNO, MNN-$\mathcal{H}^2$, and BCR-Net either rely on overlapping multi-resolution balls in a U-Net-like V-cycle or apply locally connected convolutions across scales.
In contrast, the GridNFMM explicitly isolates the non-overlapping interaction lists ($\mathcal{U}_\tau$) of the FMM downward pass via factorised message passing, strictly separating compressed far-field translations from direct near-field convolutions.

\paragraph{ML preconditioners and modifying existing neural operators}
Another line of research combines multigrid methods with convolutional and spectral networks.
Lerer et al.~\cite{lerer_multigrid-augmented_2024} modify a U-Net \cite{ronneberger_u-net_2015} by placing an FNO at the coarsest level to serve as a preconditioner for a Krylov solver on the Helmholtz equation, though the medium encodings must be recomputed when the PDE coefficient changes.
To directly improve neural operators on high-frequency wave problems, Benitez et al.~\cite{benitez_out--distributional_2023} proposed $s\text{NO}+\varepsilon I$, which separates the non-local operator $\mathcal{K}_t$ from the local transformation and incorporates stochastic depth to improve generalisation on the Helmholtz map.
More broadly, several works extend the FNO \cite{li_fourier_2021, duruisseaux_fourier_2025}: factorising the Fourier transform along problem dimensions (F-FNO \cite{tran_factorized_2023}), combining FNO with U-Net pathways (U-FNO \cite{wen_u-fno_2022}), using global convolutional kernels (CNO \cite{raonic_convolutional_2023}), or augmenting Fourier convolutions with localised differential and integral kernels in image space \cite{liu-schiaffini_neural_2024}.
To alleviate the $\mathcal{O}(T k_{\text{max}}^d c_v^2)$ parameter growth of dense spectral weight tensors, the Tucker-Factorised FNO (TFNO) \cite{kossaifi_multi-grid_2023, kossaifi2025librarylearningneuraloperators} compresses the joint weight tensor across modes, channels, and layers via a low-rank Tucker decomposition, serving as our primary parameter-matched benchmark.

\subsection{The FMM}\label{sec:fmm_background}

Originally developed by Rokhlin \cite{rokhlin_rapid_1985} and Greengard and Rokhlin \cite{greengard_fast_1987} for $N$-body gravitational and electrostatic fields, and extended to kernel-independent elliptic PDEs by Ying et al.~\cite{ying_kernel-independent_2004}, the Fast Multipole Method evaluates pairwise interactions of the form
\begin{equation}\label{eqn:n_body_sum}
  u(r_i) = \sum_{j=1}^N G(r_i, r_j) f(r_j), \qquad i = 1, \dots, N,
\end{equation}
or continuous volume potentials $u(r) = \int_{\mathcal{D}} \mathcal{E}(r, r') f(r')\,dr'$ in $\mathcal{O}(N)$ operations \cite{martinsson_fast_2019, ying_pedestrian_2012}.
Fundamentally, the classical FMM exploits the property that the free-space fundamental solution $\mathcal{E}(r, r')$ is smooth when the target point $r$ and source point $r'$ are well-separated.
This smoothness implies that the admissible off-diagonal blocks of the interaction matrix are numerically low-rank, admitting a separation of variables of truncation degree $p$, $\mathcal{E}(r, r') \approx \sum_{q=0}^p B_q(r) C_q(r')$.
Consequently, the volume potential at a target point $r \in \mathcal{D}$ can be partitioned into a singular or rapidly varying local contribution over the near field $\overline{\mathcal{N}_r}$, which is evaluated directly, and a smooth far-field integral over $\mathcal{D} \setminus \overline{\mathcal{N}_r}$, which is approximated hierarchically by reusing compressed source representations across many target evaluations.

\begin{figure}[t]
  \centering
  \includegraphics[width=0.45\textwidth]{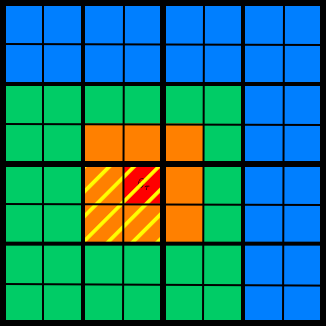}
  \caption{Hierarchical domain decomposition of $\mathcal{D}$ for a target box $B_\tau$ (red) in the 2D FMM. The domain is partitioned into the parent's far field $\mathcal{F}_{\Tau}$ (blue), containing boxes well-separated from $B_\tau$'s parent $B_\Tau$ (yellow dashed boundary); the interaction list or unique far field $\mathcal{U}_\tau$ (green), containing boxes at the same level well-separated from $B_\tau$ whose parents are adjacent to $B_\Tau$; and the near field $\mathcal{N}_\tau$ (orange), consisting of boxes directly adjacent to $B_\tau$.}
  \label{fig:decomposition_omega}
\end{figure}

To leverage this separation systematically, the FMM partitions the spatial domain $\mathcal{D}$ recursively using a quadtree in 2D (or an octree in 3D) across levels $\ell \in \{0, \dots, d_T\}$, where $\ell = 0$ denotes the root and $\ell = d_T$ denotes the leaf level containing $M_\ell = 4^\ell$ boxes in 2D.
We use subscript $\tau$ for target boxes and $\sigma$ for source boxes.
For any two boxes $B_\tau$ and $B_\sigma$ at the same level $\ell$ with centres $c_\tau, c_\sigma$ and side length $2a$, the boxes are said to be \emph{well-separated} (admissible) if $\|c_\tau - c_\sigma\|_\infty \ge 4a$, meaning they are separated by at least one buffer box of side length $2a$.
As illustrated in \cref{fig:decomposition_omega}, this admissibility criterion defines two primary interaction sets for each target box $B_\tau$.
First, the \emph{near field} $\mathcal{N}_\tau$ comprises the boxes at the same level $\ell$ directly adjacent to $B_\tau$, with $\mathcal{L}_\tau = \mathcal{N}_\tau \cup \{B_\tau\}$ denoting the $3 \times 3$ local neighbourhood of at most $9$ boxes in 2D.
Second, the \emph{interaction list} (or \emph{unique far field}) $\mathcal{U}_\tau$ comprises the boxes at level $\ell$ that are well-separated from $B_\tau$ while their parents are adjacent to (or coincide with) the parent $B_\Tau$ of $B_\tau$.
In two dimensions, $|\mathcal{U}_\tau| \le 27$, and the precomputed interaction masks defining $\mathcal{U}_\tau$ on level $\ell = 3$ of the quadtree are shown in \cref{fig:quadtree_lv3_masks}.

\begin{figure}[t]
  \centering
  \includegraphics[width=0.95\textwidth]{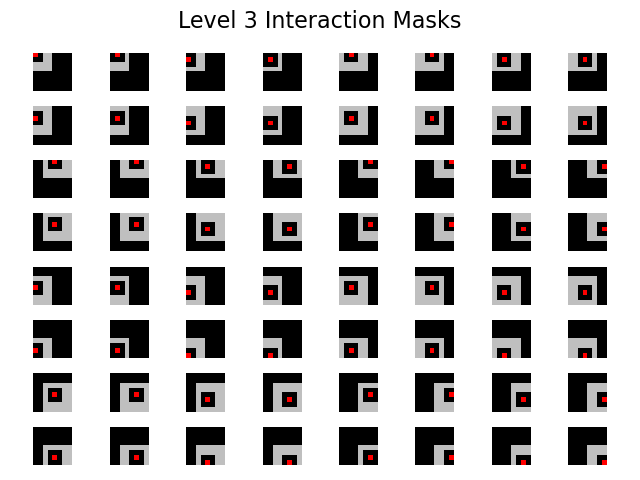}
  \caption{Precomputed interaction masks on level $\ell = 3$ of the 2D quadtree for eight representative target box positions. In each panel, the red square denotes the target box $B_\tau$, grey squares denote active source boxes in the interaction list $B_\sigma \in \mathcal{U}_\tau$, and black squares denote masked boxes (either handled at coarser levels via $\mathcal{F}_\Tau$ or deferred to the near field $\mathcal{L}_\tau$).}
  \label{fig:quadtree_lv3_masks}
\end{figure}

To achieve $\mathcal{O}(N)$ complexity, the multilevel FMM executes five linear operators across three stages, whose correspondence with standard FMM terminology is summarised in \cref{tab:fmm_nomenclature}.

\begin{table}[t]
  \centering
  \footnotesize
  \setlength{\tabcolsep}{4.5pt}
  \caption{Correspondence between standard FMM stage nomenclature and operator notation.}
  \label{tab:fmm_nomenclature}
  \begin{tabular}{lll}
    \toprule
    \textbf{Standard FMM Stage} & \textbf{Operator} & \textbf{Description} \\
    \midrule
    Particle-to-Multipole (P2M) & $\ofs{\sigma}$ & Outgoing expansion from leaf sources in $B_\sigma$ \\
    Multipole-to-Multipole (M2M) & $\ofo{\Sigma,\sigma}$ & Upward shift from child $B_\sigma$ to parent $B_\Sigma$ \\
    Multipole-to-Local (M2L) & $\ifo{\tau,\sigma}$ & Far-field translation from $B_\sigma \in \mathcal{U}_\tau$ to target $B_\tau$ \\
    Local-to-Local (L2L) & $\ifi{\tau,\Tau}$ & Downward shift from parent $B_\Tau$ to child $B_\tau$ \\
    Local-to-Particle (L2P) & $\tfi{\tau}$ & Target evaluation from incoming expansion in $B_\tau$ \\
    \bottomrule
  \end{tabular}
\end{table}

\paragraph{Upward pass}
The upward pass aggregates source information from the leaves to the coarser levels of the tree.
At the leaf level ($\ell = d_T$), sources $\{f(r_j)\}_{j \in \mathcal{I}_\sigma}$ inside leaf box $B_\sigma$ are embedded into a compact outgoing multipole coefficient vector $q_\sigma \in \mathbb{R}^{c_T}$ using the \emph{outgoing-from-sources} (P2M) operator $\ofs{\sigma}$ in \cref{eqn:OFS_definition}.
Propagating from level $\ell = d_T$ up to $\ell = 2$, the outgoing expansion $q_\Sigma$ for each parent box $B_\Sigma$ is formed by translating and summing the outgoing expansions of its four children $\sigma \in \mathcal{C}_\Sigma$ via the \emph{outgoing-from-outgoing} (M2M) shift operator $\ofo{\Sigma,\sigma}$ in \cref{eqn:OFO_definition}:
\begin{align}
  q_\sigma &= \ofs{\sigma}\big([f(r_j), r_j]_{j \in \mathcal{I}_\sigma}\big), \label{eqn:OFS_definition} \\
  q_\Sigma &= \sum_{\sigma \in \mathcal{C}_\Sigma} \ofo{\Sigma, \sigma} q_\sigma. \label{eqn:OFO_definition}
\end{align}

\paragraph{Downward pass}
The downward pass distributes far-field potentials from the coarse levels down to the leaves.
Starting at level $\ell = 2$ (since levels $0$ and $1$ are too coarse to admit well-separated boxes, so $h_\Tau = \mathbf{0}$ at $\ell = 2$) and descending to $\ell = d_T$, the incoming local expansion $h_\tau \in \mathbb{R}^{c_T}$ for target box $B_\tau$ is constructed from two components: the parent's incoming expansion $h_\Tau$ shifted to the child's centre via the \emph{incoming-from-incoming} (L2L) operator $\ifi{\tau,\Tau}$, and the newly well-separated source boxes in the interaction list $\mathcal{U}_\tau$ translated via the \emph{incoming-from-outgoing} (M2L) operator $\ifo{\tau,\sigma}$:
\begin{equation}\label{eqn:Downward_Pass_Equation}
  h_\tau = \ifi{\tau, \Tau} h_\Tau + \sum_{\sigma \in \mathcal{U}_\tau} \ifo{\tau, \sigma} q_\sigma.
\end{equation}

\paragraph{Leaf level pass}
At the finest level of refinement ($\ell = d_T$), the potential at each target point $r_i \in B_\tau$ is evaluated by expanding the accumulated incoming vector $h_\tau$ via the \emph{target-from-incoming} (L2P) operator $\tfi{\tau}$ and adding the direct near-field interactions over $\mathcal{L}_\tau = \mathcal{N}_\tau \cup \{B_\tau\}$:
\begin{equation}\label{eqn:level_Leaf_Level}
  u(r_i) = \underbrace{\sum_{B_\sigma \in \mathcal{N}_\tau \cup \{B_\tau\}} \; \sum_{r'_j \in B_\sigma} \mathcal{E}(r_i, r'_j) f(r'_j)}_{\text{Near Field (Direct)}} + \underbrace{\tfi{\tau}(r_i; h_\tau)}_{\text{Far Field (Tree)}}.
\end{equation}

% =============================================================================
\section{Method}\label{sec:method}
% =============================================================================

Our primary objective in developing a Neural Fast Multipole Method is the construction of neural network layers that implement the hierarchical tree traversal of the FMM algorithm while replacing handcrafted analytical expansions with learnable operators.
To preserve the algorithmic structure of the classical FMM while adapting it to data-driven PDE operator learning, the network layers are designed to implement three core mechanisms: they operate on the hierarchical domain decomposition defined by the FMM tree and its sparse interaction lists $\mathcal{U}_\tau$; they employ a parameter-efficient factorised approximation of the interaction kernel throughout the tree; and they combine linear or affine translation maps for the latent potentials with normalisation and non-linear gated updates to model heterogeneous PDEs.

\begin{table}[ht!]
  \centering
  \footnotesize
  \caption{Summary of GridNFMM architecture hyperparameters and notation.}
  \label{tab:GridNFMM_dimension_notation}
  \begin{tabular}{ll}
    \toprule
    \textbf{Notation} & \textbf{Description} \\
    \midrule
    $c_L$   & Leaf-level feature dimension \\
    $c_T$   & Tree-level latent multipole/local dimension \\
    $c_H$   & Pointwise channel-mixer (GLU / lifting / projection) hidden dimension \\
    $d_T$   & Depth of the FMM quadtree ($M = 4^{d_T}$ leaf boxes) \\
    \bottomrule
  \end{tabular}
\end{table}

To this end, we introduce the \textbf{Grid-based Neural Fast Multipole Method (GridNFMM)}, whose tree translation layers model the FMM equations \cref{eqn:OFS_definition,eqn:OFO_definition,eqn:Downward_Pass_Equation,eqn:level_Leaf_Level} within a neural operator architecture.
Because the purpose of the GridNFMM is to establish the learnable tree-level architecture of the FMM, we make the simplifying structural assumption that there is at most one point within each leaf-level box ($P = 1$).
For a uniform 2D grid of $N$ points, a quadtree of depth $d_T$ partitions the domain $\mathcal{D}$ into $M = 4^{d_T} = N$ disjoint leaf boxes (for example, $64 \times 64 = 4096$ leaf boxes at depth $d_T = 6$).
While this assumption ties the spatial resolution of the GridNFMM to the depth of the quadtree, preventing the model from transferring zero-shot across discretisations, it enables all tree shifts and leaf evaluations to be expressed cleanly on a uniform hierarchical grid.

Tessellating the domain $\mathcal{D}$ into $M = 4^{d_T}$ disjoint leaf boxes as outlined in \cref{sec:fmm_background} corresponds to reshaping the row-stacked data matrix of $N$ grid points, $\mathbf{X} \in \mathbb{R}^{N \times c_{\text{in}}}$ (or $\mathbb{R}^{B_b \times N \times c_{\text{in}}}$ across a batch of size $B_b$), with $c_{\text{in}}$ input channels $[r_i^\top, f(r_i)^\top]$, into $\mathbf{X} \in \mathbb{R}^{M \times c_{\text{in}}}$ across $M = N$ leaf boxes containing one point each.
In the GridNFMM, we index the $M_\ell = 4^\ell$ boxes at each tree level $\ell \in \{1, \dots, d_T\}$ using a 1D Morton ($Z$-order) space-filling curve.
Morton ordering linearises the 2D spatial hierarchy so that the four children $\mathcal{C}_\Sigma$ of any parent box $B_\Sigma$ at level $\ell - 1$ always occupy four contiguous indices at level $\ell$.
This contiguity enables upward $4:1$ coarsening ($\ofo{}$) and downward $1:4$ refinement ($\ifi{}$) to be executed as zero-copy tensor reshapes while preserving the parent--child containment relations of the quadtree.

\subsection{GridNFMM block}\label{sec:GridNFMM_Block}

The GridNFMM embeds a stack of $T$ GridNFMM blocks within the decoupled Neural Operator architecture \cref{eqn:Neural_Operator_Unrolled,eqn:New_Neural_Operator_Step}, using a 2-layer SiLU MLP for the pointwise lifting operator $\mathbf{P}$ and a pre-LayerNorm, spectrally normalised 2-layer SiLU MLP for the pointwise projection operator $\mathbf{Q}$.
Within each block, the translation maps are modelled on the linear operators of the FMM and combined following its bidirectional information flow, while LayerNorm's and the gated DNN update step makes the complete block non-linear.
Whereas standard neural operators such as the FNO employ a single global integral kernel for the spatial mixer $\mathcal{K}_t$, the GridNFMM block decomposes spatial mixing into complementary near-field and far-field pathways: a local $3 \times 3$ CNN for direct neighbourhood interactions, and a hierarchical quadtree pass for long-range potential evaluation.

\begin{figure}[ht!]
  \centering
  \includegraphics[width=0.56\textwidth]{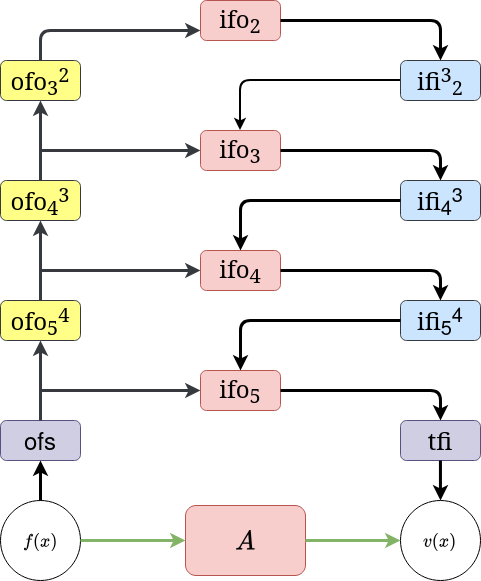}
  \caption{Schematic architecture of the GridNFMM block, illustrated at tree depth $d_T = 5$. Purple nodes denote leaf encoding and decoding operators ($\ofs{}, \tfi{}$); gold and blue nodes indicate upward ($\ofo{}$) and downward ($\ifi{}$) shift operators, respectively; red nodes represent interaction-list far-field translation ($\ifo{}$) across $\mathcal{U}_\tau$; and the green pathway denotes near-field direct evaluation via a factorised $3 \times 3$ CNN ($\mathcal{A}$) with gated fusion and a residual GLU update.}
  \label{fig:GridNFMM_Architecture}
\end{figure}

The schematic architecture of the GridNFMM block is illustrated in \cref{fig:GridNFMM_Architecture}.
We start with the leaf feature vector $x_\sigma \in \mathbb{R}^{c_L}$ (representing the lifted field at leaf box $B_\sigma$) as the input, which is mapped into an outgoing latent multipole vector $q_\sigma \in \mathbb{R}^{c_T}$ by $\ofs{}$.
This outgoing representation is then propagated up the levels of the tree by $\ofo{\Sigma, \sigma}$.
On each level within the tree, the incoming latent local vector $h_\tau \in \mathbb{R}^{c_T}$ is computed via $\ifo{\tau, \sigma}$ applied to $q_\sigma$ for each box in the interaction list $B_\sigma \in \mathcal{U}_\tau$, combined with the parent's incoming representation $h_\Tau$ shifted via $\ifi{\tau, \Tau}$.
At the finest level, the incoming representations are decoded into the far-field feature vector $y_\tau \in \mathbb{R}^{c_L}$ for leaf box $B_\tau$ by $\tfi{}$, while the near-field contribution is computed directly by the local spatial mixer $\mathcal{A}_\theta$.
In our implementation, linear projections in $\ofs{}$, $\tfi{}$, the factorised $\ifo{}$ value and output maps, the leaf CNN, and the final projection $\mathbf{Q}$ are wrapped with spectral normalisation \cite{miyato_spectral_2018} to stabilise training.
The block is governed by four structural hyperparameters, summarised in \cref{tab:GridNFMM_dimension_notation}.

\subsection{Box encoding and decoding}\label{sec:grid_ofs_tfi}

The box encoding and decoding operators correspond to $\ofs{}$ and $\tfi{}$, coloured purple in \cref{fig:GridNFMM_Architecture}.
Under the single-point-per-leaf assumption of the GridNFMM, each grid point is situated at the centre of its leaf box $B_\sigma$, so the relative displacement from the point to the box centre is identically zero across all leaves.
Consequently, $\ofs{}$ and $\tfi{}$ share the same weight matrices across all $M$ leaf boxes, preserving translation equivariance across the uniform leaf grid.

\subsubsection{Outgoing-from-Sources (OFS)}\label{sec:grid_ofs_operator}

The Outgoing-from-Sources operator $\ofs{} : \mathbb{R}^{c_L} \to \mathbb{R}^{c_T}$ maps the leaf feature vector $x_\sigma \in \mathbb{R}^{c_L}$ within each leaf box $B_\sigma$ into a box-level latent multipole coefficient vector $q_\sigma \in \mathbb{R}^{c_T}$:
\begin{equation}\label{eqn:grid_ofs_operator}
  q_\sigma = \mathbf{W}_V^\bullet \operatorname{LayerNorm}_{\mathrm{in}}(x_\sigma) = \mathbf{W}_V \operatorname{LayerNorm}_{\mathrm{in}}(x_\sigma) + b_{\text{ofs}} \in \mathbb{R}^{c_T},
\end{equation}
where $\mathbf{W}_V \in \mathbb{R}^{c_T \times c_L}$ and $b_{\text{ofs}} \in \mathbb{R}^{c_T}$ form a learnable affine projection $\mathbf{W}_V^\bullet$ shared across all leaf boxes $\sigma \in \{1, \dots, M\}$.
For the row-stacked leaf feature matrix $\mathbf{X} \in \mathbb{R}^{M \times c_L}$, this acts pointwise as $\mathbf{Q}_L = \operatorname{LayerNorm}_{\mathrm{in}}(\mathbf{X})\mathbf{W}_V^\top + \mathbf{1}_M b_{\text{ofs}}^\top \in \mathbb{R}^{M \times c_T}$, encoding each leaf's feature vector into the tree's latent multipole space.

\subsubsection{Target-from-Incoming (TFI)}\label{sec:grid_tfi_operator}

The Target-from-Incoming operator $\tfi{} : \mathbb{R}^{c_T} \to \mathbb{R}^{c_L}$ is constructed symmetrically to $\ofs{}$ in \cref{eqn:grid_ofs_operator}, decoding the accumulated box-level latent local coefficient vector $h_\tau \in \mathbb{R}^{c_T}$ at the end of the downward pass back into the leaf feature space:
\begin{equation}\label{eqn:grid_tfi_operator}
  y_\tau = \mathbf{W}_V^\bullet h_\tau = \mathbf{W}_V h_\tau + b_{\text{tfi}} \in \mathbb{R}^{c_L},
\end{equation}
where $\mathbf{W}_V \in \mathbb{R}^{c_L \times c_T}$ and $b_{\text{tfi}} \in \mathbb{R}^{c_L}$ form a learnable affine projection $\mathbf{W}_V^\bullet$ shared across all leaf boxes $\tau \in \{1, \dots, M\}$ (or $\mathbf{Y} = \mathbf{H}_L \mathbf{W}_V^\top + \mathbf{1}_M b_{\text{tfi}}^\top \in \mathbb{R}^{M \times c_L}$ in row-stacked form).
This parametrisation of $\tfi{}$ is affine in $h_\tau$ and translation-equivariant across the leaf grid.

\subsection{Shift operators}\label{sec:grid_shifts}

The FMM algorithm shifts compressed representations between adjacent levels $\ell \in \{2, \dots, d_T\}$ of the tree via the upward (coarsening) shift operator $\ofo{}$ and the downward (refinement) shift operator $\ifi{}$.
Mathematically, these act as linear operators on the latent multipole and local coefficient vectors within the tree.
In the classical FMM for the Laplacian, the $\ofo{}$ (M2M) and $\ifi{}$ (L2L) shift operators take a triangular form under polynomial or multipole bases.
In the learnable GridNFMM setting, triangular weight matrices mirror this sparsity: the upward map is upper-triangular and the downward map is lower-triangular, while Rotary Position Embeddings (RoPE) \cite{su_roformer_2023} distinguish the relative quadrant offsets of the four children in the downward map.

\subsubsection{Outgoing-from-Outgoing (OFO) operator}\label{sec:grid_ofo_operator}

At tree level $\ell \in \{3, \dots, d_T\}$, the Outgoing-from-Outgoing operator $\ofo{\Sigma, \sigma} : \mathbb{R}^{c_T} \to \mathbb{R}^{c_T}$ computes the latent multipole vector $q_\Sigma \in \mathbb{R}^{c_T}$ of a parent box $B_\Sigma$ (at level $\ell - 1$) by aggregating the latent multipole vectors $\{q_\sigma\}_{\sigma \in \mathcal{C}_\Sigma}$ of its four children $\mathcal{C}_\Sigma = \{\sigma_1, \dots, \sigma_4\}$:
\begin{equation}\label{eqn:grid_ofo_operator}
  q_\Sigma = \sum_{\sigma \in \mathcal{C}_\Sigma} \mathbf{W}_U q_\sigma \in \mathbb{R}^{c_T},
\end{equation}
where $\mathbf{W}_U \in \mathbb{R}^{c_T \times c_T}$ is a learnable, bias-free upper-triangular transformation shared across all boxes at level $\ell$, with distinct parameters per tree level.
The upward map sums the four transformed child vectors, omitting a bias prevents its accumulation across tree levels during coarsening.

\subsubsection{Incoming-from-Incoming (IFI) operator}\label{sec:grid_ifi_operator}

At tree level $\ell \in \{3, \dots, d_T\}$, the Incoming-from-Incoming operator $\ifi{\tau, \Tau} : \mathbb{R}^{c_T} \to \mathbb{R}^{c_T}$ shifts the latent local vector $h_\Tau \in \mathbb{R}^{c_T}$ of a parent box $B_\Tau$ (at level $\ell - 1$) down to each of its four children $B_\tau$ ($\tau \in \mathcal{C}_\Tau$ at level $\ell$):
\begin{equation}\label{eqn:grid_ifi_operator}
  \tilde{h}_\tau = \mathbf{W}_D \mathcal{R}_\tau (h_\Tau) \in \mathbb{R}^{c_T},
\end{equation}
where $\mathbf{W}_D \in \mathbb{R}^{c_T \times c_T}$ is a learnable, bias-free lower-triangular weight matrix shared across all boxes at level $\ell$, and $\mathcal{R}_\tau \in \mathbb{R}^{c_T \times c_T}$ is the RoPE rotation matrix encoding the child's quadrant index within $B_\Tau$.
Here $\tilde{h}_\tau \in \mathbb{R}^{c_T}$ denotes the shifted parent local coefficient vector distributed to child box $B_\tau$.

\subsection{Incoming-from-Outgoing (\texorpdfstring{$\ifo{}$}{T-ifo}) operator}\label{sec:grid_ifo_operator}

At each tree level $\ell \in \{2, \dots, d_T\}$, the Incoming-from-Outgoing operator $\ifo{\tau, \sigma} : \mathbb{R}^{c_T} \to \mathbb{R}^{c_T}$ translates an aggregated latent multipole vector $q_\sigma \in \mathbb{R}^{c_T}$ from a well-separated source box $B_\sigma \in \mathcal{U}_\tau$ into a latent local contribution centred at target box $B_\tau$.

\subsubsection{Motivation and MPNN formulation}\label{sec:grid_ifo_motivation}

The motivation for the design of the learnable $\ifo{\tau, \sigma}$ operator is to establish a formulation that is parameter-efficient, expressive, and extendable from 2D to 3D domains.
On a uniform hierarchical grid, the relative displacement vectors $c_\tau - c_\sigma$ between interacting boxes at level $\ell$ are restricted to at most $K$ distinct relative offsets, as summarised in Remark~\ref{rmk:num_unique_ifo_operators}.

\begin{remark}[Unique incoming-from-outgoing translation operators]\label{rmk:num_unique_ifo_operators}
  In a domain discretised by a uniform hierarchical quadtree, the relative displacement vectors $c_\tau - c_\sigma$ are highly constrained.
  At any given level, each target box has at most $27$ interaction-list boxes in 2D, whose relative centre displacements, measured in units of the level-$\ell$ box side length, range over the $K = 7^2 - 3^2 = 40$ vectors $\{(i, j) \in \mathbb{Z}^2 : 2 \le \max(|i|, |j|) \le 3\}$ (indexed within a $7^2 = 49$ offset table).
  Consequently, for a two-dimensional uniform quadtree, at most $40$ distinct $\ifo{}$ matrices are required to evaluate all far-field interactions across an entire tree level (in a 3D octree, each box has $|\mathcal{U}_\tau| \le 189$ interaction-list neighbours with offsets among $K = 7^3 - 3^3 = 316$ vectors).
\end{remark}

Interpreting the $\ifo{\tau, \sigma}$ stage within the Message-Passing Neural Network (MPNN) framework \cite{gilmer_neural_2017} on the sparse interaction graph $\mathcal{E}_\ell = \{(\tau, \sigma) : B_\sigma \in \mathcal{U}_\tau\}$, each directed edge $(\tau, \sigma)$ carries a relative displacement offset indexed by $\delta(\tau, \sigma) \in \{1, \dots, K\}$.
Consequently, the FMM downward pass \cref{eqn:Downward_Pass_Equation} at level $\ell$ decomposes directly into the MPNN message, aggregation, and update operations in \cref{tab:downward_pass_mpnn}.

\begin{table}[ht!]
  \centering
  \footnotesize
  \caption{Downward-pass decomposition in the MPNN framework.}
  \label{tab:downward_pass_mpnn}
  \begin{tabular}{ll}
    \toprule
    \textbf{Name} & \textbf{Function} \\
    \midrule
    Message Function & $m_{\tau, \sigma} = \ifo{\tau, \sigma} q_\sigma$ \\
    \addlinespace[2pt]
    Aggregation Function & $m_\tau = \sum_{\sigma \in \mathcal{U}_\tau} m_{\tau, \sigma}$ \\
    \addlinespace[2pt]
    Update Function & $h_\tau = m_\tau + \tilde{h}_\tau$ \\
    \bottomrule
  \end{tabular}
\end{table}

While translation invariance across a uniform tree level restricts the number of distinct operators per level to $K$, learning $K$ dense matrices in $\mathbb{R}^{c_T \times c_T}$ at every tree level $\ell \in \{2, \dots, d_T\}$ incurs $\mathcal{O}(K c_T^2)$ parameters per level, which becomes prohibitive in 3D.
To reduce this parameter count, we factorise $\ifo{\tau, \sigma}$ across the sparse interaction graph defined by $\mathcal{U}_\tau$.

\subsubsection{Factorised matrix parametrisation}\label{sec:grid_ifo_factorised}

Rather than learning $K$ dense matrices per level, we factorise $\ifo{\tau, \sigma}$ into shared input and output channel-mixing matrices and $K$ diagonal scaling matrices corresponding to the $K$ distinct relative displacement offsets at level $\ell$.
Let $\mathbf{\Theta} \in \mathbb{R}^{K \times c_T}$ store the learnable diagonal entries at level $\ell$, and let $\delta : \mathcal{E}_\ell \to \{1, 2, \dots, K\}$ map each interaction-list pair $(\tau, \sigma)$ ($\sigma \in \mathcal{U}_\tau$) to its relative displacement index $k = \delta(\tau, \sigma)$, with diagonal matrix $\mathbf{\Theta}_k = \operatorname{diag}(\mathbf{\Theta}_{k,:}) \in \mathbb{R}^{c_T \times c_T}$.
Because the uniform quadtree geometry is static, the index lookup $\delta(\tau, \sigma)$ depends only on the relative box coordinates and is precomputed at initialisation.
We then define $\ifo{\tau, \sigma} : \mathbb{R}^{c_T} \to \mathbb{R}^{c_T}$ at level $\ell$ using two bias-free dense matrices $\mathbf{W}_V, \mathbf{W}_O \in \mathbb{R}^{c_T \times c_T}$ (shared across all boxes and edges at level $\ell$) and the offset-specific diagonal matrix $\mathbf{\Theta}_{\delta(\tau, \sigma)} \in \mathbb{R}^{c_T \times c_T}$:
\begin{equation}\label{eqn:ifo_matrix_decomp}
  \ifo{\tau, \sigma} = \mathbf{W}_O \mathbf{\Theta}_{\delta(\tau, \sigma)} \mathbf{W}_V \in \mathbb{R}^{c_T \times c_T}.
\end{equation}
While \cref{eqn:ifo_matrix_decomp} resembles a singular value decomposition (SVD) with shared left and right bases, $\mathbf{W}_V$ and $\mathbf{W}_O$ are left unconstrained and the diagonal entries of $\mathbf{\Theta}_{\delta(\tau, \sigma)}$ are signed real numbers, reducing the per-level parameter count from $\mathcal{O}(K c_T^2)$ to $\mathcal{O}(c_T^2 + K c_T)$.
For each target box $B_\tau$ at level $\ell$, the shared dense projections $\mathbf{W}_V$ and $\mathbf{W}_O$ are applied only once per box, while the interaction-list contribution $m_\tau \in \mathbb{R}^{c_T}$ is accumulated over $\mathcal{U}_\tau$ via $\mathcal{O}(c_T)$ elementwise products:
\begin{align}\label{eqn:ifo_scatter_gather}
  z_\tau &= \sum_{\sigma \in \mathcal{U}_\tau} \mathbf{\Theta}_{\delta(\tau, \sigma)}\mathbf{W}_V q_\sigma \in \mathbb{R}^{c_T}, \nonumber \\
  m_\tau &= \mathbf{W}_O z_\tau = \sum_{\sigma \in \mathcal{U}_\tau}\ifo{\tau,\sigma}q_\sigma \in \mathbb{R}^{c_T}.
\end{align}
The total incoming local coefficient vector $h_\tau \in \mathbb{R}^{c_T}$ at box $B_\tau$ combines $m_\tau$ with the shifted parent local vector $\tilde{h}_\tau = \ifi{\tau, \Tau} h_\Tau$ via a gated update:
\begin{equation}\label{eqn:ifo_gating}
  h_\tau = m_\tau + \left( \mathbf{1}_{c_T} + \mathbf{W}_G^\bullet(z_\tau) \right) \odot \tilde{h}_\tau \in \mathbb{R}^{c_T},
\end{equation}
where $\mathbf{W}_G^\bullet(z_\tau) = \mathbf{W}_G z_\tau + b_G$ is a learnable affine gating map with weight $\mathbf{W}_G \in \mathbb{R}^{c_T \times c_T}$ and bias $b_G \in \mathbb{R}^{c_T}$ initialised to $\mathbf{0}$, ensuring that \cref{eqn:ifo_gating} initialises to the classical linear superposition $h_\tau = m_\tau + \tilde{h}_\tau$ \cref{eqn:Downward_Pass_Equation}, while allowing learned multiplicative modulation across tree levels during training.

\subsection{Leaf level}\label{sec:grid_leaf}

At the leaf level ($\ell = d_T$), the local spatial mixer $\mathcal{A}_\theta$ approximates the direct near-field interactions \cref{eqn:level_Leaf_Level}.
Because each leaf box contains a single grid point on a uniform 2D Cartesian grid, the local neighbourhood $\mathcal{L}_\tau = \mathcal{N}_\tau \cup \{B_\tau\}$ forms a $3 \times 3$ kernel of $9$ adjacent leaf boxes.
We therefore parametrise $\mathcal{A}_\theta$ as a factorised $3 \times 3$ convolutional kernel shared across all leaf boxes.

Specifically, let $\hat{x}_\sigma = \operatorname{LayerNorm}_{\mathrm{in}}(x_\sigma)$ denote the input-normalised features shared by the tree and leaf branches.
For each target leaf box $B_\tau$, the operator gathers $\{\hat{x}_\sigma\}_{\sigma \in \mathcal{L}_\tau}$, with zero-padding outside the domain boundary, and combines an isotropic local average with an anisotropic low-rank correction over the $9$ ordered neighbours $\sigma_1, \dots, \sigma_9 \in \mathcal{L}_\tau$:
\begin{align}\label{eqn:grid_leaf_cnn}
  \bar{x}_\tau &= \frac{1}{9} \sum_{\sigma \in \mathcal{L}_\tau} \hat{x}_\sigma \in \mathbb{R}^{c_L}, \nonumber \\
  a_\tau &= \mathbf{W}_O^\bullet \left( \bar{x}_\tau + \mathbf{W}_{\text{mix}}^\bullet \left( \bigoplus_{\sigma \in \mathcal{L}_\tau} \mathbf{W}_{\text{feat}} \operatorname{LayerNorm}_{\mathrm{CNN}}(\hat{x}_\sigma) \right) \right) \in \mathbb{R}^{c_L},
\end{align}
where $\mathbf{W}_{\text{feat}} \in \mathbb{R}^{r \times c_L}$ is a bias-free linear projection compressing each normalised neighbour vector to width $r = \max(1, \lfloor c_L / 4 \rfloor)$, $\bigoplus_{\sigma \in \mathcal{L}_\tau}$ concatenates the $9$ compressed vectors in fixed kernel order into $\mathbb{R}^{9r}$, $\mathbf{W}_{\text{mix}}^\bullet : \mathbb{R}^{9r} \to \mathbb{R}^{c_L}$ is an affine spatial mixer carrying a learnable bias, and $\mathbf{W}_O^\bullet : \mathbb{R}^{c_L} \to \mathbb{R}^{c_L}$ is an affine output channel mixer with bias.
The CNN has its own LayerNorm, distinct from the shared input normalisation; padded neighbours are zeroed after this second normalisation.

We fuse the near-field feature vector $a_\tau \in \mathbb{R}^{c_L}$ with the decoded far-field feature vector $y_\tau \in \mathbb{R}^{c_L}$ from \cref{eqn:grid_tfi_operator} using a state-dependent SiLU gate, followed by a pre-LayerNorm residual GLU channel mixer:
\begin{align}
  g_\tau &= \operatorname{SiLU} \left( \mathbf{W}_g^\bullet \left( x_\tau + a_\tau  \right) \right) \in \mathbb{R}^{c_L}, \label{eqn:grid_leaf_gate} \\
  c_\tau &= x_\tau + a_\tau + g_\tau \odot y_\tau \in \mathbb{R}^{c_L}, \label{eqn:grid_leaf_combine} \\
  x_\tau^{t+1} &= c_\tau + \operatorname{GLU} \left( \operatorname{LayerNorm}(c_\tau) \right) \in \mathbb{R}^{c_L}, \label{eqn:grid_leaf_update}
\end{align}
where $\mathbf{W}_g^\bullet(x) = \mathbf{W}_g x + b_g$ is an affine map whose weight and bias are initialised to zero, giving $g_\tau = 0$ at initialisation.
Within the decoupled neural operator block structure \cref{eqn:New_Neural_Operator_Step}, the sum $a_\tau + g_\tau \odot y_\tau$ constitutes the spatial mixer $\mathcal{K}_t$ combining local interactions ($a_\tau$) and hierarchical far-field interactions ($y_\tau$), while $\operatorname{GLU}(\operatorname{LayerNorm}(\cdot))$ serves as the pointwise channel mixer $\operatorname{DNN}_t$.

\subsection{Algorithm and computational complexity}\label{sec:gridnfmm-complexity}

We now make the forward computation of the \emph{GridNFMM block} explicit and state its asymptotic computational cost.
We consider a domain discretised by $N$ points, sorted into the leaves of a balanced quadtree of depth $L = d_T = \log_4 N$, so that the number of boxes at level $\ell$ is $M_\ell = 4^\ell$ and the leaf level satisfies $M_L = N$.
Let $c = \max(c_L, c_T, c_H)$ denote the maximum channel width, which we treat as a constant with respect to $N$.
\Cref{alg:gridnfmm} summarises one forward pass of the GridNFMM block.

\begin{algorithm}[ht!]
  \caption{GridNFMM Block}
  \label{alg:gridnfmm}
  \begin{algorithmic}[1]
    \REQUIRE Leaf feature matrix $\mathbf{X} \in \mathbb{R}^{B \times N \times c_L}$, sparse interaction lists $\{\mathcal{E}_\ell\}_{\ell=2}^{L}$, near-field interaction list $\mathcal{N}_{\text{near}}$, leaf mask $m_L$ (all ones on a full grid)
    \ENSURE Updated leaf feature matrix $\mathbf{U} \in \mathbb{R}^{B \times N \times c_L}$
    \STATE $\hat{\mathbf{X}} \leftarrow \operatorname{LayerNorm}(\mathbf{X})$
    \STATE \COMMENT{Upward pass: P2M \eqref{eqn:grid_ofs_operator} and M2M \eqref{eqn:grid_ofo_operator}}
    \STATE $\mathbf{Q}_L \leftarrow (\hat{\mathbf{X}}\mathbf{W}_V^\top + \mathbf{1}_N b_{\mathrm{ofs}}^\top)\odot m_L$ \hfill\COMMENT{P2M: encode to tree multipole vectors}
    \FOR{$\ell = L, L-1, \dots, 3$}
      \STATE $\mathbf{Q}_{\ell-1} \leftarrow \operatorname{OFO}_\ell(\mathbf{Q}_\ell)$ \hfill\COMMENT{M2M: 4:1 child-to-parent aggregation}
      \STATE $m_{\ell-1} \leftarrow \operatorname{PoolMask}(m_\ell)$ \hfill\COMMENT{parent valid if any child valid}
      \STATE $\mathbf{Q}_{\ell-1} \leftarrow \mathbf{Q}_{\ell-1} \odot m_{\ell-1}$
    \ENDFOR
    \STATE \COMMENT{Downward pass: M2L \eqref{eqn:ifo_scatter_gather}--\eqref{eqn:ifo_gating} and L2L \eqref{eqn:grid_ifi_operator}}
    \STATE $\mathbf{H}_2 \leftarrow \operatorname{IFO}_2(\mathbf{Q}_2,\, \mathbf{0};\, \mathcal{E}_2)$ \hfill\COMMENT{M2L: interaction list translation}
    \FOR{$\ell = 3, 4, \dots, L$}
      \STATE $\tilde{\mathbf{H}}_\ell \leftarrow \operatorname{IFI}_{\ell-1}(\mathbf{H}_{\ell-1}) \odot m_\ell$ \hfill\COMMENT{L2L: parent-to-child shift}
      \STATE $\mathbf{H}_\ell \leftarrow \operatorname{IFO}_\ell(\mathbf{Q}_\ell,\, \tilde{\mathbf{H}}_\ell;\, \mathcal{E}_\ell) \odot m_\ell$ \hfill\COMMENT{M2L: far-field interaction and gate}
    \ENDFOR
    \STATE \COMMENT{Leaf level: L2P \eqref{eqn:grid_tfi_operator}, near-field CNN \eqref{eqn:grid_leaf_cnn}, and update \eqref{eqn:grid_leaf_gate}--\eqref{eqn:grid_leaf_update}}
    \STATE $\mathbf{Y} \leftarrow \operatorname{TFI}(\mathbf{H}_L)$ \hfill\COMMENT{L2P: decode far field to $\mathbb{R}^{B \times N \times c_L}$}
    \STATE $\mathbf{A} \leftarrow \operatorname{CNN}_{3\times 3}(\hat{\mathbf{X}};\, \mathcal{N}_{\text{near}})$ \hfill\COMMENT{3$\times$3 near-field CNN}
    \STATE $\mathbf{G} \leftarrow \operatorname{SiLU}\bigl(\mathbf{W}_g^\bullet(\mathbf{X} + \mathbf{A})\bigr)$ \hfill\COMMENT{zero-initialised affine gate}
    \STATE $\mathbf{C} \leftarrow \mathbf{X} + \mathbf{A} + \mathbf{G} \odot \mathbf{Y}$
    \STATE $\mathbf{U} \leftarrow \mathbf{C} + \operatorname{GLU}\bigl(\operatorname{LayerNorm}(\mathbf{C})\bigr)$ \hfill\COMMENT{GLU residual update}
    \RETURN $\mathbf{U}$
  \end{algorithmic}
\end{algorithm}

Although the block performs a full bidirectional tree traversal followed by a local convolution, each of its constituent operators acts across a single box, point, or interaction edge at a time.

\begin{theorem}[Complexity of the GridNFMM block]\label{thm:gridnfmm-complexity}
  Let the $N$ grid points occupy the leaves of a balanced quadtree of depth $L = d_T = \log_4 N$ (with $1$ point per leaf box), with sparse interaction lists of bounded size $|\mathcal{U}_\tau| \le 27$ per box, and let $c = \max(c_L, c_T, c_H)$.
  Then one forward pass of the GridNFMM block requires $\mathcal{O}\!\left(N c^2\right)$ floating-point operations and $\mathcal{O}\!\left(N c\right)$ activation and index memory.
  In particular, for $c$ bounded independently of $N$, the block runs in $\mathcal{O}(N)$ time.
\end{theorem}

The complete proof of \cref{thm:gridnfmm-complexity} is provided in \cref{apdx:complexity_proof}.

\FloatBarrier
% =============================================================================
\section{Numerical experiments}\label{sec:experiments}
% =============================================================================

\subsection{Loss functions and metrics}\label{sec:loss_and_metrics}

To train the GridNFMM and TFNO models and evaluate both global and local prediction quality across problem scales, we employ a regularised relative $L_2$ training objective alongside three validation metrics.

\paragraph{Relative $L_2$ loss and metric}
The primary training objective is a regularised relative $L_2$ loss. For a ground-truth target $u$ and predicted output $\hat{u}$ evaluated over $N$ grid points $\{r_i\}_{i=1}^N$, the per-sample loss is computed as:
\begin{equation}\label{eqn:rel_l2_loss}
  \mathcal{L}_{\mathrm{rel}}(u, \hat{u}) = \frac{\sqrt{\sum_{i=1}^N \|u(r_i) - \hat{u}(r_i)\|_2^2 + \epsilon}}{\sqrt{\sum_{i=1}^N \|u(r_i)\|_2^2 + \epsilon} + \epsilon}, \qquad \epsilon = 10^{-8},
\end{equation}
where the channel norm includes all predicted components and $\epsilon$ stabilises gradients near zero error or zero target norm.
The reported relative $L_2$ error metric ($\mathcal{E}_2^{\text{rel}}$ or $L_2^{\text{rel}}$) evaluates the unregularised per-sample ratio averaged over the validation set:
\begin{equation}\label{eqn:rel_l2_metric}
  \mathcal{E}_2^{\text{rel}}(u, \hat{u}) = \frac{\sqrt{\sum_{i=1}^N \|u(r_i) - \hat{u}(r_i)\|_2^2}}{\sqrt{\sum_{i=1}^N \|u(r_i)\|_2^2}}.
\end{equation}
This scale-invariant metric normalises the error by the magnitude of the ground-truth solution, enabling consistent comparison across different PDE regimes.

\paragraph{Root Mean Squared Error (RMSE)}
To measure absolute average accuracy across the grid, we also report the root mean squared error:
\begin{equation}\label{eqn:rmse_metric}
  \operatorname{RMSE}(u, \hat{u}) = \sqrt{\frac{1}{N} \sum_{i=1}^N \|u(r_i) - \hat{u}(r_i)\|_2^2}.
\end{equation}

\paragraph{$L_\infty$ norm}
The $L_\infty$ error captures the worst-case pointwise absolute discrepancy between the predicted and ground-truth fields:
\begin{equation}\label{eqn:linf_metric}
  \|u - \hat{u}\|_\infty = \max_{1 \le i \le N} \|u(r_i) - \hat{u}(r_i)\|_\infty.
\end{equation}
In our evaluation harness, for sample $s$ with pointwise maximum $e_s = \max_{i, c} |\hat{u}_{s,c}(r_i) - u_{s,c}(r_i)|$, the logged $L_\infty$ statistic aggregates physical validation batches $b$ of size $B_b = 8$ as $E_\infty^{\mathrm{batch}} = (\sum_b B_b \max_{s \in b} e_s)/(\sum_b B_b)$ across identical validation batches.
Unlike averaged metrics such as RMSE, $L_\infty$ ensures that large localised errors (such as near sharp scatterer interfaces) are not masked by smooth performance elsewhere in the domain.

\subsection{Experimental setup}\label{sec:experimental_setup}

We evaluate the GridNFMM against the Tucker-Factorised Fourier Neural Operator (TFNO) \cite{kossaifi_multi-grid_2023, kossaifi2025librarylearningneuraloperators} on the 2D PDE benchmark problems outlined in Kovachki et al.~\cite{kovachki_neural_2021} and available in the \texttt{neuraloperator} Python library \cite{kossaifi2025librarylearningneuraloperators}, as well as on custom datasets generated from the 2D Helmholtz equation defined over a heterogeneous medium.
Because the spatial resolution of the GridNFMM is tied to the depth of the quadtree, all training and validation examples are discretised on a $64 \times 64$ uniform grid ($N = 4096$), corresponding to a tree depth of $d_T = 6$.

\begin{table}[ht!]
  \centering
  \footnotesize
  \renewcommand{\arraystretch}{1.15}
  \caption{Architectural hyperparameters and computational footprint on an NVIDIA A100 GPU ($64 \times 64$ grid, batch size $8$). ``Memory Size'' denotes peak resident model state (trainable parameters plus precomputed sparse interaction index tensors and persistent buffers).}
  \label{tab:GridNFMM_hyperparameters}
  \begin{tabular}{l c c}
    \toprule
    \textbf{Hyperparameter} & \textbf{TFNO} & \textbf{GridNFMM} \\
    \midrule
    Number of Blocks ($T$) & 4 & 4 \\
    Hidden Dimension ($c_H$) & 256 & 256 \\
    Feature / Tree Dimension ($c_L, c_T$) & 64 & 64 \\
    Fourier Modes / Tree Depth & 12 (Modes, $\text{rank}=0.5$) & 6 (Depth $d_T$) \\
    \cmidrule(lr){1-3}
    Trainable Parameters & 886,713 & 844,353 \\
    Memory Size & 6.12 MB & 86.25 MB \\
    Speed (sec/batch) & $4.18 \times 10^{-2}$ s & $8.06 \times 10^{-2}$ s \\
    \bottomrule
  \end{tabular}
\end{table}

\paragraph{Model hyperparameters and training information}
To ensure a fair comparison, both models are constrained to similar trainable parameter counts ($844,353$ for GridNFMM and $886,713$ for TFNO with Tucker $\text{rank} = 0.5$) and employ identical training hyperparameters (\cref{tab:GridNFMM_hyperparameters}).
Specifically, both architectures use a 2-layer MLP for the lifting and projection operators ($\mathbf{P}, \mathbf{Q}$), substituting only the spatial mixing operators $\mathcal{K}_t$ within the decoupled Neural Operator block structure \cref{eqn:New_Neural_Operator_Step}.
For every model and dataset, we train for $50$ epochs using the AdamW optimiser ($\beta_1 = 0.9, \beta_2 = 0.95$) with decoupled weight decay of $10^{-5}$, a batch size of $8$, and the regularised relative $L_2$ loss \cref{eqn:rel_l2_loss}.
The learning rate increases linearly from zero to $10^{-3}$ over a warm-up period of $5$ epochs, followed by cosine decay to zero over the remaining $45$ epochs.
All reported validation metrics reflect single training runs with fixed random seeds.

\paragraph{Helmholtz dataset}
We focus on the 2D inhomogeneous Helmholtz equation defined over a heterogeneous medium:
\begin{equation}\label{eqn:helmholtz_pde}
  \mathcal{L}_n[u^s] \equiv \Delta u^s(r) + k_0^2 n(r) u^s(r) = (1 - n(r)) k_0^2 u^i(r), \qquad r \in \mathcal{D},
\end{equation}
which breaks down into three operator-learning cases: in Case~1, the material profile $n(r)$ is held constant while the incident wavefield $u^i(r)$ varies; in Case~2, the material profile $n(r)$ varies while the incident wavefield $u^i(r)$ remains fixed as a point source ($k_0 = 8$); and in Case~3, both $n(r)$ and $u^i(r)$ vary simultaneously ($k_0 = 8$).
Within these cases, further physical distinctions are made based on the contrast between the background wavenumber $k_0$ and the scattering medium $n(r)$: \emph{Weak Scattering}, where $0.9 k_0 \le n(r) \le 1.1 k_0$; \emph{Transmission}, where $k_0 \le n(r) \le 2 k_0$; and \emph{Hard Scattering}, where $2 k_0 \ll n(r)$.
We note that due to the $64 \times 64$ grid resolution, for high wavenumbers inside dense scatterers the spatial wavefield approaches the Nyquist sampling limit, which can cause pointwise $L_\infty$ errors to peak inside the scattering objects.
Full details of the finite-element dataset generation in NGSolve are provided in \cref{apdx:data_generation}.

\subsection{Results}\label{sec:results}

\Cref{tab:GridNFMM_ex_results_table,tab:GridNFMM_case1_results_table,tab:GridNFMM_case2_results_table,tab:GridNFMM_case3_results_table} report the validation error metrics (RMSE, $\mathcal{E}_2^{\text{rel}} / L_2^{\text{rel}}$, and $L_\infty$) comparing the GridNFMM and the TFNO across the external benchmarks and the three Helmholtz cases.

\paragraph{External benchmarks}
We begin by comparing the GridNFMM and the TFNO on the 2D Darcy Flow ($a(r) \mapsto u(r)$) and Vorticity Navier--Stokes ($w_0(r) \mapsto w(r)$) datasets from the \texttt{neuraloperator} library \cite{kovachki_neural_2021, kossaifi2025librarylearningneuraloperators} in \cref{tab:GridNFMM_ex_results_table}.
As expected from the elliptic structure of the FMM, the GridNFMM outperforms the TFNO on the elliptic Darcy Flow benchmark across all metrics (reducing relative $L_2$ error from $0.0650$ to $0.0294$ and RMSE from $0.0390$ to $0.0179$), whereas the TFNO achieves lower error on the non-linear Vorticity Navier--Stokes problem ($\mathcal{E}_2^{\text{rel}} = 0.0794$ vs.\ $0.139$).
Qualitative field predictions and absolute error maps for the Navier--Stokes dataset are shown in \cref{fig:comparison_shared_target}.

\begin{table}[ht!]
  \centering
  \footnotesize
  \renewcommand{\arraystretch}{1.2}
  \caption{External datasets: Comparison of the GridNFMM to the TFNO on a $64 \times 64$ grid.}
  \label{tab:GridNFMM_ex_results_table}
  \begin{tabular}{ll ccc ccc}
    \toprule
    \multicolumn{2}{c}{\textbf{PDE Problem}} & \multicolumn{3}{c}{\textbf{GridNFMM}} & \multicolumn{3}{c}{\textbf{TFNO}} \\
    \cmidrule(lr){1-2} \cmidrule(lr){3-5} \cmidrule(lr){6-8}
    \textbf{Problem Type} & \textbf{Input} & RMSE & $\mathcal{E}_{2}^{\text{rel}}$ & $L_\infty$ & RMSE & $\mathcal{E}_{2}^{\text{rel}}$ & $L_\infty$\\
    \midrule
    Darcy Flow & $a(r)$ & \textbf{0.0179} & \textbf{0.0294} & \textbf{0.158} & 0.0390 & 0.0650 & 0.299 \\
    Vorticity Navier--Stokes & $w_0(r)$ & 0.149 & 0.139 & 1.25 & \textbf{0.0841} & \textbf{0.0794} & \textbf{0.473} \\
    \bottomrule
  \end{tabular}
\end{table}

\begin{figure}[ht!]
  \centering
  \begin{minipage}[c]{0.26\textwidth}
    \centering
    \includegraphics[width=\linewidth]{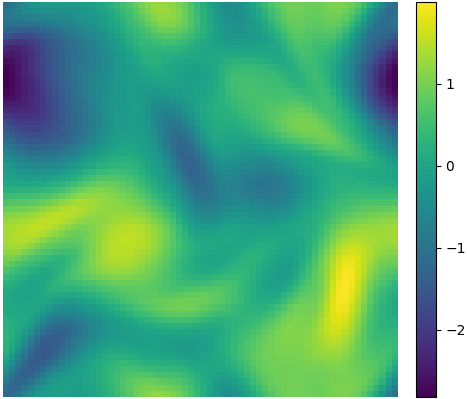}\\[0.2em]
    {\footnotesize (a) Target}
  \end{minipage}%
  \hfill\vrule\hfill%
  \begin{minipage}[c]{0.68\textwidth}
    \begin{minipage}[b]{0.48\linewidth}
      \centering
      \includegraphics[width=\linewidth]{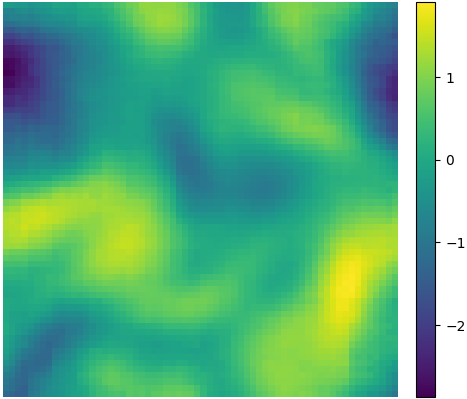}\\[0.2em]
      {\footnotesize (b) GridNFMM Prediction}
    \end{minipage}\hfill
    \begin{minipage}[b]{0.48\linewidth}
      \centering
      \includegraphics[width=\linewidth]{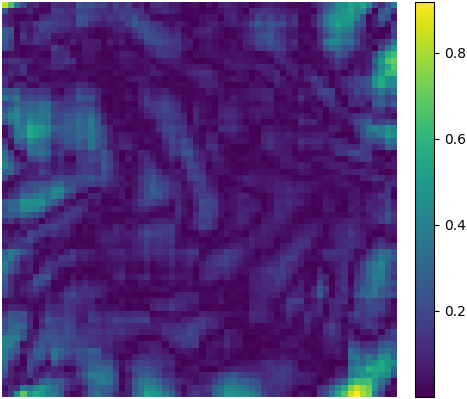}\\[0.2em]
      {\footnotesize (c) GridNFMM Abs.\ Error}
    \end{minipage}\\[0.5em]
    \begin{minipage}[b]{0.48\linewidth}
      \centering
      \includegraphics[width=\linewidth]{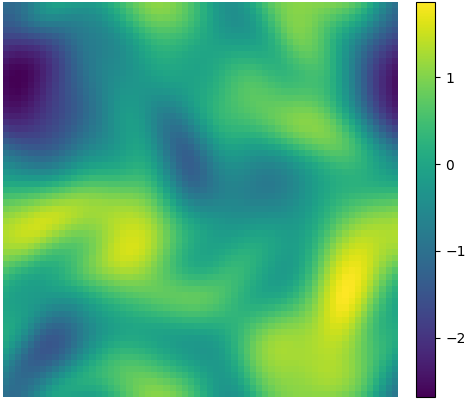}\\[0.2em]
      {\footnotesize (d) TFNO Prediction}
    \end{minipage}\hfill
    \begin{minipage}[b]{0.48\linewidth}
      \centering
      \includegraphics[width=\linewidth]{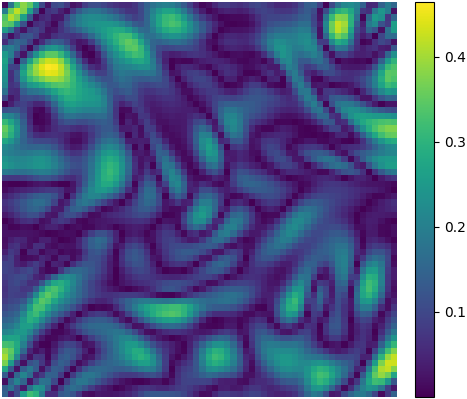}\\[0.2em]
      {\footnotesize (e) TFNO Abs.\ Error}
    \end{minipage}
  \end{minipage}
  \caption{Comparison against ground-truth target on 2D Vorticity Navier--Stokes: Target (a), GridNFMM prediction and absolute error (b, c), and TFNO prediction and absolute error (d, e).}
  \label{fig:comparison_shared_target}
\end{figure}

\paragraph{Case 1}
In Case~1, we solve \cref{eqn:helmholtz_pde} with a fixed material profile $n(r)$ and a variable incident wave $u^i(r)$.
Because the medium $n(r)$ is fixed, Case~1 corresponds to parameterising the linear Green's integral operator with a neural network architecture:
\begin{equation}\label{eqn:case1_integral}
  u^s(r) = \int_{\mathcal{D}} \mathcal{S}_{n}(r, r') u^i(r') \, dr',
\end{equation}
where $\mathcal{S}_n(r, r')$ denotes the Green's kernel for the fixed heterogeneous medium $n(r)$.
As shown in \cref{tab:GridNFMM_case1_results_table}, the GridNFMM outperforms the TFNO across all four configurations and all three error metrics, reducing the relative $L_2$ error on Hard Scattering from $0.286$ to $0.0986$ and more than halving the $L_\infty$ error on high-$k_0$ plane-wave scattering ($0.333$ vs.\ $0.770$).
Qualitative field predictions for Case~1 ($k_0 = 12$ plane wave) are presented in \cref{fig:case1_transmission_plane_k0_12_2x3}.

\begin{table}[ht!]
  \centering
  \footnotesize
  \setlength{\tabcolsep}{4.5pt}
  \renewcommand{\arraystretch}{1.15}
  \caption{Case 1: Comparison of the GridNFMM to the TFNO on a $64 \times 64$ grid (validation set).}
  \label{tab:GridNFMM_case1_results_table}
  \begin{tabular}{l c c ccc ccc}
    \toprule
    \multicolumn{2}{c}{\textbf{Case 1}} & \textbf{Wavenumber} & \multicolumn{3}{c}{\textbf{GridNFMM}} & \multicolumn{3}{c}{\textbf{TFNO}} \\
    \cmidrule(lr){1-2} \cmidrule(lr){3-3} \cmidrule(lr){4-6} \cmidrule(lr){7-9}
    \textbf{Problem Type} & \textbf{Source} & $k_0$ & RMSE & $L_2^{\text{rel}}$ & $L_\infty$ & RMSE & $L_2^{\text{rel}}$ & $L_\infty$ \\
    \midrule
    High $k_0$ & Point & 12 & \textbf{0.0389} & \textbf{0.123} & \textbf{0.275} & 0.0695 & 0.220 & 0.442 \\
    High $k_0$ & Plane & 12 & \textbf{0.0604} & \textbf{0.261} & \textbf{0.333} & 0.107 & 0.464 & 0.770 \\
    Multiple Sources & Both & 8 & \textbf{0.0303} & \textbf{0.111} & \textbf{0.160} & 0.0638 & 0.235 & 0.388 \\
    Hard Scattering & Both & 8 & \textbf{0.0244} & \textbf{0.0986} & \textbf{0.191} & 0.0705 & 0.286 & 0.519 \\
    \bottomrule
  \end{tabular}
\end{table}

\begin{figure}[ht!]
  \centering
  \setlength{\tabcolsep}{4pt}
  \begin{tabular}{@{} c c c @{}}
    \textbf{Input and Target} & \textbf{Prediction} & \textbf{Absolute Error} \\[0.3em]
    \includegraphics[width=0.315\linewidth]{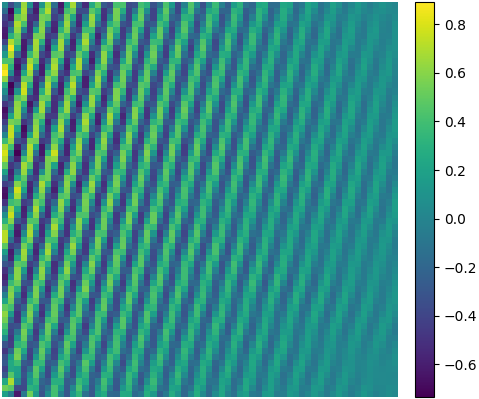} &
    \includegraphics[width=0.315\linewidth]{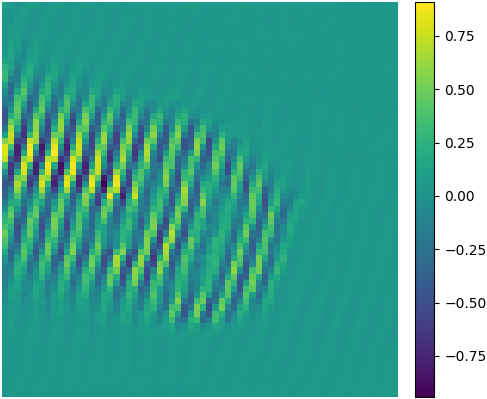} &
    \includegraphics[width=0.315\linewidth]{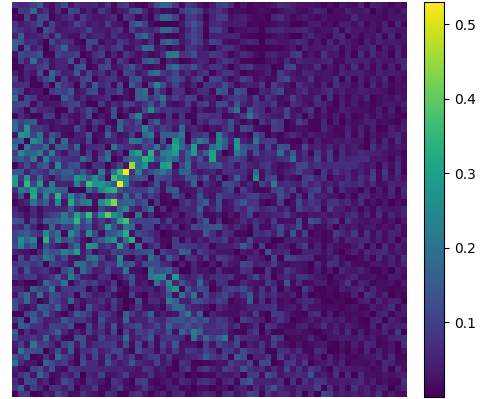} \\
    \footnotesize Input $u^i(r)$ & \footnotesize GridNFMM Prediction & \footnotesize GridNFMM Abs.\ Error \\[0.5em]
    \includegraphics[width=0.315\linewidth]{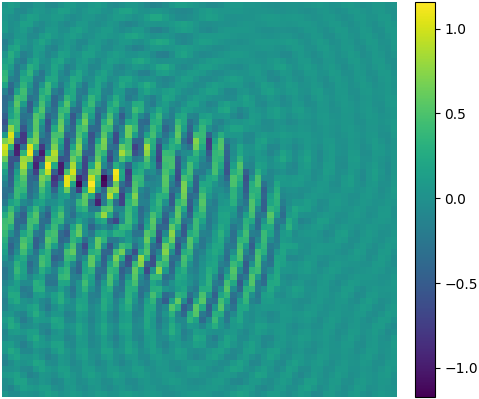} &
    \includegraphics[width=0.315\linewidth]{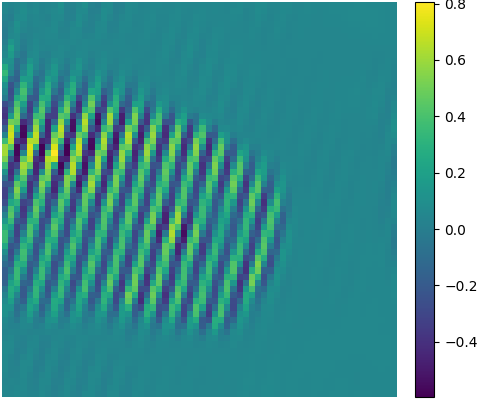} &
    \includegraphics[width=0.315\linewidth]{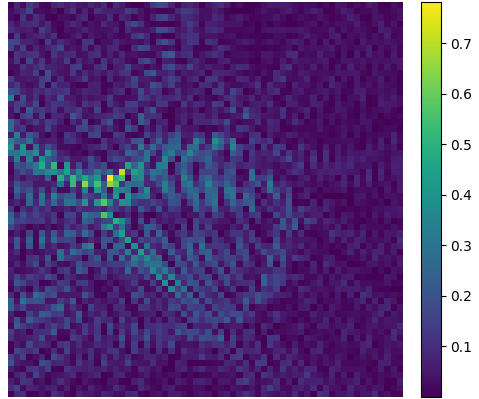} \\
    \footnotesize Target $u^s(r)$ & \footnotesize TFNO Prediction & \footnotesize TFNO Abs.\ Error
  \end{tabular}
  \caption{Comparison on \texttt{Case1\_Transmission\_Plane\_k0\_12}: Shared input and target (left column), GridNFMM prediction and absolute error (top right), and TFNO prediction and absolute error (bottom right).}
  \label{fig:case1_transmission_plane_k0_12_2x3}
\end{figure}

\paragraph{Case 2}
In Case~2, we solve \cref{eqn:helmholtz_pde} for a variable medium profile $n(r)$ and a fixed point-source incident field $u^i(r)$ at background wavenumber $k_0 = 8$.
\Cref{tab:GridNFMM_case2_results_table} reports the validation errors across Weak Scattering, Transmission, and Hard Scattering, while \cref{fig:validation_metrics} shows the validation convergence curves on \texttt{Case2\_Hard\_Scattering\_k0\_8} and \cref{fig:case2_hard_scattering_k0_8_2x3} displays a representative qualitative comparison.
The GridNFMM outperforms the TFNO across all three regimes and metrics, achieving $L_2^{\text{rel}} = 0.0703$ and $L_\infty = 0.495$ on Hard Scattering compared with $0.0972$ and $0.545$ for the TFNO.

\begin{table}[ht!]
  \centering
  \footnotesize
  \renewcommand{\arraystretch}{1.15}
  \caption{Case 2: Comparison of the GridNFMM to the TFNO on a $64 \times 64$ grid (validation set).}
  \label{tab:GridNFMM_case2_results_table}
  \begin{tabular}{l c ccc ccc}
    \toprule
    \multicolumn{2}{c}{\textbf{Case 2}} & \multicolumn{3}{c}{\textbf{GridNFMM}} & \multicolumn{3}{c}{\textbf{TFNO}} \\
    \cmidrule(lr){1-2} \cmidrule(lr){3-5} \cmidrule(lr){6-8}
    \textbf{Problem Type} & \textbf{Input} & RMSE & $L_2^{\text{rel}}$ & $L_\infty$ & RMSE & $L_2^{\text{rel}}$ & $L_\infty$ \\
    \midrule
    Weak Scattering & $n(r)$ & \textbf{0.00790} & \textbf{0.0377} & \textbf{0.0618} & 0.00797 & 0.0381 & 0.0661 \\
    Transmission & $n(r)$ & \textbf{0.0465} & \textbf{0.145} & \textbf{0.328} & 0.0480 & 0.150 & 0.346 \\
    Hard Scattering & $n(r)$ & \textbf{0.0237} & \textbf{0.0703} & \textbf{0.495} & 0.0311 & 0.0972 & 0.545 \\
    \bottomrule
  \end{tabular}
\end{table}

\begin{figure}[ht!]
  \centering
  \begin{minipage}[t]{0.32\textwidth}
    \centering
    \includegraphics[width=\linewidth]{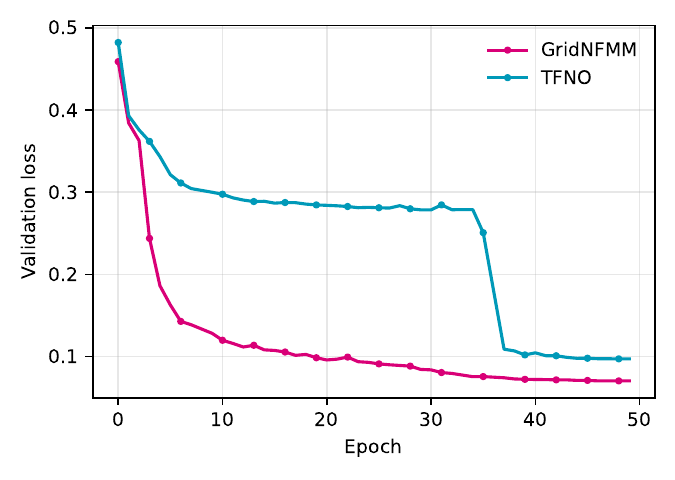}\\[0.2em]
    {\footnotesize (a) Validation loss}
  \end{minipage}\hfill
  \begin{minipage}[t]{0.32\textwidth}
    \centering
    \includegraphics[width=\linewidth]{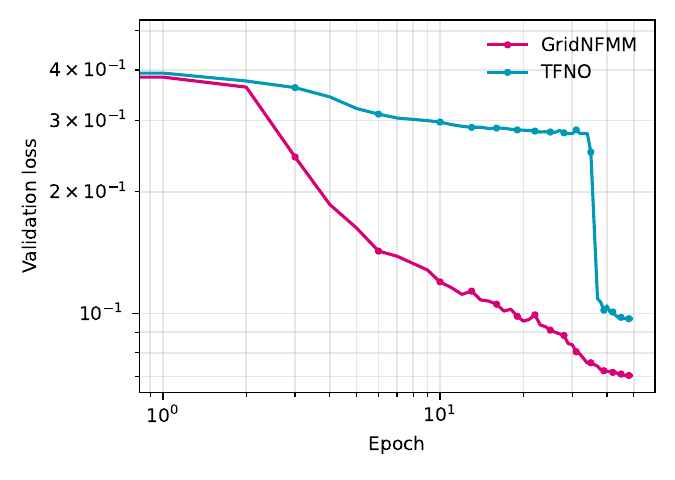}\\[0.2em]
    {\footnotesize (b) Log--log scale}
  \end{minipage}\hfill
  \begin{minipage}[t]{0.32\textwidth}
    \centering
    \includegraphics[width=\linewidth]{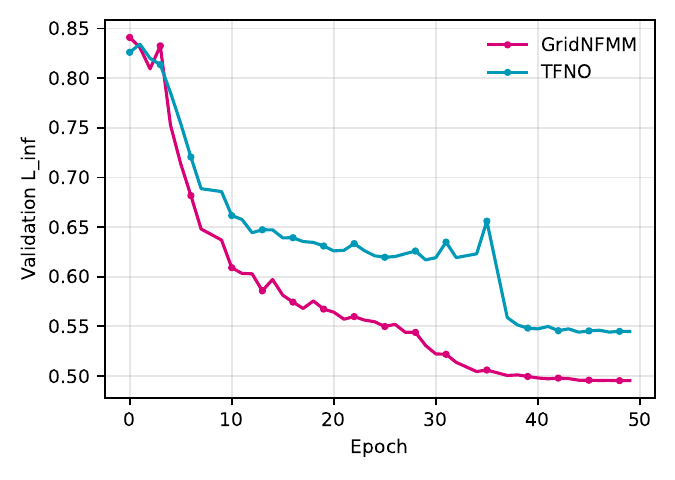}\\[0.2em]
    {\footnotesize (c) $L_\infty$ error}
  \end{minipage}
  \caption{Loss curves on the \texttt{Case2\_Hard\_Scattering\_k0\_8} dataset for the GridNFMM and TFNO: (a) linear validation loss, (b) log--log scale, and (c) $L_\infty$ error.}
  \label{fig:validation_metrics}
\end{figure}

\begin{figure}[ht!]
  \centering
  \setlength{\tabcolsep}{4pt}
  \begin{tabular}{@{} c c c @{}}
    \textbf{Input and Target} & \textbf{Prediction} & \textbf{Absolute Error} \\[0.3em]
    \includegraphics[width=0.315\linewidth]{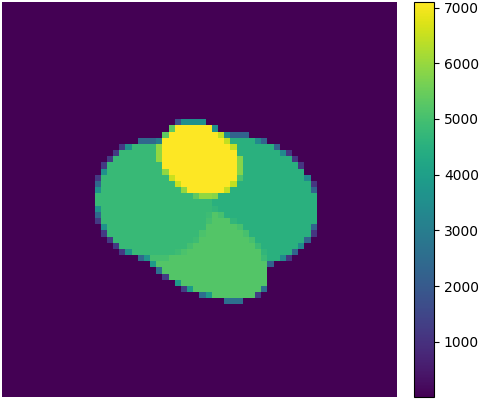} &
    \includegraphics[width=0.315\linewidth]{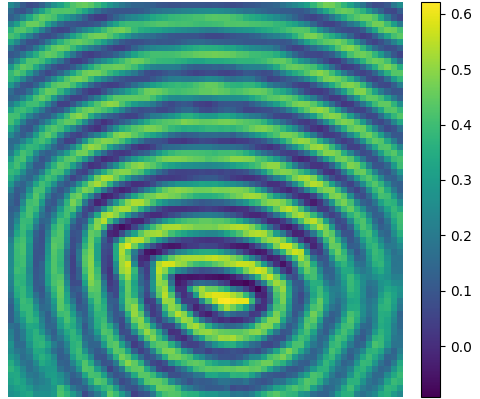} &
    \includegraphics[width=0.315\linewidth]{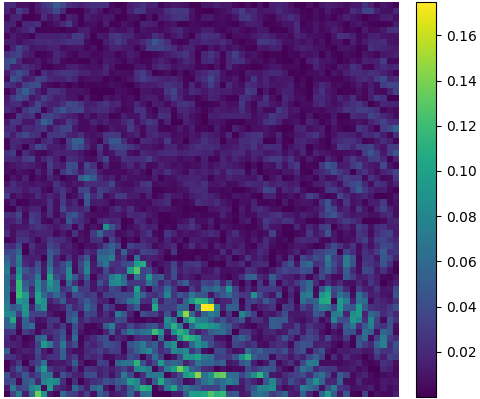} \\
    \footnotesize Input $n(r)$ & \footnotesize GridNFMM Prediction & \footnotesize GridNFMM Abs.\ Error \\[0.5em]
    \includegraphics[width=0.315\linewidth]{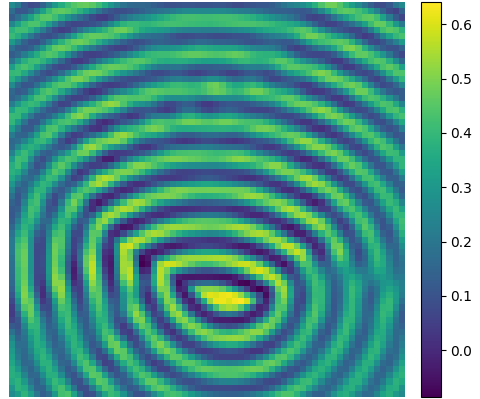} &
    \includegraphics[width=0.315\linewidth]{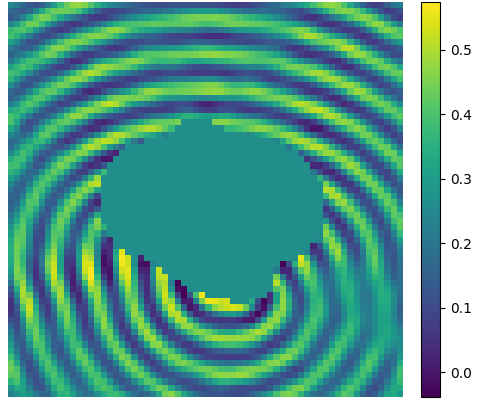} &
    \includegraphics[width=0.315\linewidth]{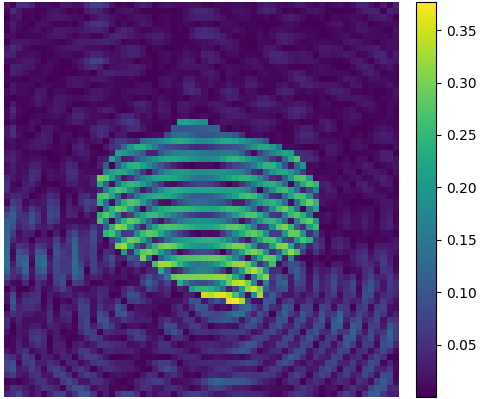} \\
    \footnotesize Target $u^s(r)$ & \footnotesize TFNO Prediction & \footnotesize TFNO Abs.\ Error
  \end{tabular}
  \caption{Comparison on \texttt{Case2\_Hard\_Scattering\_k0\_8}: Shared input and target (left column), GridNFMM prediction and absolute error (top right), and TFNO prediction and absolute error (bottom right).}
  \label{fig:case2_hard_scattering_k0_8_2x3}
\end{figure}

\paragraph{Case 3}
In Case~3, we evaluate the forward problem of predicting the scattered field $u^s(r)$ when both the medium $n(r)$ and the incident wave $u^i(r)$ vary simultaneously, providing the combined forcing $f(r) = (1 - n(r)) k_0^2 u^i(r)$ as input.
As reported in \cref{tab:GridNFMM_case3_results_table}, the GridNFMM again outperforms the TFNO across all three scattering regimes and metrics, reducing the relative $L_2$ error on Hard Scattering from $0.270$ to $0.166$ and on Weak Scattering from $0.231$ to $0.152$.
Representative field predictions and error profiles for Case~3 are provided in \cref{apdx:visualisations} (\cref{fig:case3_transmission_k0_8_2x3}).

\begin{table}[ht!]
  \centering
  \footnotesize
  \renewcommand{\arraystretch}{1.15}
  \caption{Case 3: Comparison of the GridNFMM to the TFNO on a $64 \times 64$ grid (validation set).}
  \label{tab:GridNFMM_case3_results_table}
  \begin{tabular}{l c ccc ccc}
    \toprule
    \multicolumn{2}{c}{\textbf{Case 3}} & \multicolumn{3}{c}{\textbf{GridNFMM}} & \multicolumn{3}{c}{\textbf{TFNO}} \\
    \cmidrule(lr){1-2} \cmidrule(lr){3-5} \cmidrule(lr){6-8}
    \textbf{Problem Type} & \textbf{Source} & RMSE & $L_2^{\text{rel}}$ & $L_\infty$ & RMSE & $L_2^{\text{rel}}$ & $L_\infty$ \\
    \midrule
    Weak Scattering & Both & \textbf{0.0190} & \textbf{0.152} & \textbf{0.118} & 0.0294 & 0.231 & 0.194 \\
    Transmission & Both & \textbf{0.0551} & \textbf{0.188} & \textbf{0.397} & 0.0569 & 0.195 & 0.421 \\
    Hard Scattering & Both & \textbf{0.0444} & \textbf{0.166} & \textbf{0.386} & 0.0714 & 0.270 & 0.555 \\
    \bottomrule
  \end{tabular}
\end{table}

\subsection{Analysis of results}\label{sec:analysis_of_results}

\paragraph{Performance on linear vs.\ non-linear operators}
While the Darcy Flow and Helmholtz PDEs are linear in the unknown solution $u$, the parameter-to-solution mapping $a(r) \mapsto u(r)$ (for Darcy Flow) and the medium-to-solution mapping $n(r) \mapsto u^s(r)$ (for Helmholtz Cases~2 and 3) are mathematically non-linear operator problems.
As observed in \cref{tab:GridNFMM_ex_results_table}, the GridNFMM significantly outperforms the TFNO on the Darcy Flow dataset, reducing the RMSE and relative $L_2$ error by more than a factor of two, whereas the TFNO proves superior on the Vorticity Navier--Stokes dataset.
The strength of the TFNO on Navier--Stokes reflects the suitability of global Fourier modes for periodic fluid flow on equispaced grids.
Conversely, the GridNFMM is designed to approximate Green's integral representations and employs a fixed $3 \times 3$ convolution kernel at the leaf level.
Nonetheless, its substantial improvement over the TFNO on the parametric Darcy Flow benchmark and across Cases~2 and 3 of the Helmholtz benchmark demonstrates that the multipole-inspired inductive bias of the GridNFMM—augmented with non-linear gating and residual GLU channel mixing—does not restrict it to strictly linear source-to-solution operators.

\paragraph{Performance on oscillatory problems}
Across all ten custom Helmholtz configurations in \cref{tab:GridNFMM_case1_results_table,tab:GridNFMM_case2_results_table,tab:GridNFMM_case3_results_table}, the GridNFMM consistently outperforms the TFNO on every reported metric.
In Case~1 (\cref{tab:GridNFMM_case1_results_table}), where the operator is linear for a fixed medium, the GridNFMM surpasses the TFNO even at high wavenumbers ($k_0 = 12$) and under hard scattering.
In Case~2 (\cref{tab:GridNFMM_case2_results_table}) and Case~3 (\cref{tab:GridNFMM_case3_results_table}), where the medium varies and the operator is non-linear, the GridNFMM retains a clear advantage across weak scattering, transmission, and hard scattering regimes.
This indicates that combining an uncompressed local $3 \times 3$ near-field pathway with hierarchical far-field message passing over $\mathcal{U}_\tau$ resolves oscillatory wavefields and sharp scatterer interfaces more accurately than the bandwidth-limited truncated spectral convolutions of the TFNO.

\paragraph{Computational trade-offs}
\Cref{tab:GridNFMM_hyperparameters} highlights an important operational consideration: although trainable parameter counts are tightly matched ($844$k vs.\ $887$k) and the GridNFMM block scales asymptotically as $\mathcal{O}(N c^2)$ versus $\mathcal{O}(N c \log N + N c^2)$ for the TFNO, the GridNFMM requires more resident memory ($86.25$ MB vs.\ $6.12$ MB) and incurs roughly $1.9\times$ higher wall-clock latency per batch ($8.06 \times 10^{-2}$ s vs.\ $4.18 \times 10^{-2}$ s) on an NVIDIA A100 GPU.
This expanded footprint reflects the explicit tree-based routing, precomputed sparse interaction index buffers, and Python-level tree traversal overhead across levels in PyTorch, in contrast to the hardware-optimised cuFFT and Tucker kernels underpinning the TFNO.

\FloatBarrier
% =============================================================================
\section{Discussion}\label{sec:discussion}
% =============================================================================

\paragraph{Limitations and next steps}
Despite the strong empirical performance of the GridNFMM on elliptic and oscillatory PDEs, it carries a primary structural limitation: because of the simplifying assumption that there is at most one point per leaf-level box ($P = 1$), its spatial grid resolution $N = 4^{d_T}$ is tied directly to the depth $d_T$ of the quadtree ($32 \times 32$ pixels for $d_T = 5$, $64 \times 64$ for $d_T = 6$, and $128 \times 128$ for $d_T = 7$).
Consequently, the GridNFMM does not satisfy the discretisation-invariance property of a continuum Neural Operator, as its learned level-dependent parameters cannot be transferred zero-shot across different resolutions or applied directly to irregular, unstructured point clouds.
Furthermore, parametrising $\ofs{}$ and $\tfi{}$ as shared pointwise affine projections and $\mathcal{A}_\theta$ as a $3 \times 3$ CNN relies on the uniform Cartesian leaf geometry.
Relaxing the single-point-per-leaf assumption requires redesigning the leaf-level operators to handle continuous, variable point geometries ($P \ge 1$ points per leaf box) by replacing the leaf CNN and pointwise projections with coordinate-conditioned, distance-informed attention mechanisms while preserving the hierarchical tree operators ($\ofo{}, \ifo{}, \ifi{}$) developed in this work.

\paragraph{Conclusion}
In this work, we introduced the Grid-based Neural Fast Multipole Method (GridNFMM), a hierarchical neural network architecture built on the principle \emph{fix the computational graph, learn the computation}.
By replacing handcrafted, kernel-dependent multipole translation operators with learnable factorised message-passing layers across the FMM interaction lists $\mathcal{U}_\tau$, triangular inter-level shifts, and a local $3 \times 3$ convolutional pathway, the GridNFMM learns Green's and parametric PDE solution operators directly from data without requiring analytical knowledge of the interaction kernel, while maintaining $\mathcal{O}(N)$ computational complexity per block.
Our experimental evaluations demonstrate that the GridNFMM outperforms the parameter-matched Tucker-Factorised Fourier Neural Operator (TFNO) across 2D Darcy Flow and all three heterogeneous Helmholtz scattering cases, highlighting the effectiveness of embedding classical numerical solver topologies into modern machine learning architectures.

% =============================================================================
\appendix
\section{Data generation}\label{apdx:data_generation}
% =============================================================================

To generate our datasets for each Case and subcase of the 2D heterogeneous Helmholtz benchmark, we used a Finite Element Method (FEM) solver implemented in Python via NGSolve \cite{schoeberl_c11_2014}, which is built upon the automatic mesh generator NETGEN \cite{schoberl_netgen_1997}.
Below we summarise the parametric generation of the scattering field $n(r)$, the variational formulation solved in NGSolve, and the interpolation onto the uniform $64 \times 64$ grid.

\subsection{Parametric generation of the scattering field}\label{apdx:parametric_scattering}

The generation of the scattering field $n(r)$ is accomplished parametrically using elliptical scattering obstacles inspired by the Shepp--Logan phantom \cite{shepp_fourier_1974}.
Each ellipse is uniquely defined by four geometric parameters drawn independently: its centre coordinates $(x_c, y_c) \sim \mathcal{U}([-1, 1]^2)$, major semi-axis length $a \sim \mathcal{U}([0.25, 2.5])$, minor semi-axis length $b \sim \mathcal{U}([0.25, a])$, and orientation angle $\theta \sim \mathcal{U}([0, 2\pi))$.
The material properties of the ellipses are assigned relative to the background wavenumber $k_0$ according to the physical subcase: Weak Scattering ($0.9 k_0 \le n(r) \le 1.1 k_0$), Transmission ($k_0 \le n(r) \le 2 k_0$), and Hard Scattering ($2 k_0 \ll n(r)$).
To prevent a single large ellipse from obscuring smaller inclusions when multiple ellipses are present, we place candidate centres at the vertices of a regular polygon circumscribing the unit circle, with each subsequent ellipse scaled down slightly to create a natural hierarchy in the scattering field.

Depending on the operator-learning case being generated, the ellipses are sampled differently.
For Case~1, we fix the ellipses across the dataset and thus reuse the same finite-element mesh for every example while varying only the incident wave source type and location.
For Case~2, we construct nine exemplar scattering objects composed of $\{2, \dots, 10\}$ ellipses, and weakly permute these exemplar ellipses by uniformly perturbing their centre coordinates, rotation angle, and major and minor axis lengths by $\pm 10\%$ for each sample.
For Case~3, we fix the number of ellipses within the scatterer while varying every ellipse parameter freely across samples.

\subsection{Generating our datasets}\label{apdx:solver_setup}

The forward pass computes the scattered field $u^s(r)$ given an incident wavefield $u^i(r)$ and the scattering field $n(r)$.
The incident field $u^i(r)$ is generated by solving the inhomogeneous Helmholtz equation in free space:
\begin{equation}\label{eqn:apdx_incident}
  \Delta u^i(r) + k_0^2 u^i(r) = d(r),
\end{equation}
where the source term $d(r)$ is constructed as a superposition of point sources positioned along a circle outside the scatterer support or a plane wave incident at an angle perpendicular to the circle.
In Case~1, the number of distinct source positions along the circle is chosen relative to the $64 \times 64$ target grid resolution so that each rasterised incident wave in the training and validation sets is distinct ($\approx 300$ training and $100$ validation examples per source type).

Given the incident field $u^i(r)$, we solve the inhomogeneous Helmholtz equation over the variable material $n(r)$ on the bounded domain $\Omega$:
\begin{equation}\label{eqn:apdx_helmholtz_strong}
  \Delta u^s(r) + k_0^2 n(r) u^s(r) = \left(1 - n(r)\right) k_0^2 u^i(r) = f(r).
\end{equation}
Multiplying \cref{eqn:apdx_helmholtz_strong} by an arbitrary test function $v \in H^1(\Omega)$ and integrating by parts with a first-order absorbing impedance condition on the outer boundary $\Gamma_\Omega = \partial \Omega$ yields the weak formulation assembled in NGSolve:
\begin{equation}\label{eqn:apdx_helmholtz_weak}
  \int_\Omega \nabla u^s \cdot \nabla v \, dr - k_0^2 \int_\Omega n(r) u^s v \, dr - i k_0 \int_{\Gamma_\Omega} u^s v \, ds = k_0^2 \int_\Omega \left(1 - n(r)\right) u^i v \, dr.
\end{equation}
We discretise $H^1(\Omega)$ using complex-valued finite elements of polynomial order $2$, bounding the maximum mesh element size relative to $k_0$ to maintain approximately $10$ nodal points per wavelength as recommended for oscillatory FEM problems \cite{christodoulou_high-order_2017}.
Finally, because the GridNFMM operates on uniform Cartesian grids, we use NGSolve's native interpolation routine to evaluate the FEM fields onto the uniform $64 \times 64$ grid.

% =============================================================================
\section{Additional model visualisations}\label{apdx:visualisations}
% =============================================================================

\Cref{fig:case3_transmission_k0_8_2x3} presents qualitative field predictions and pointwise absolute error maps on the Case~3 Transmission dataset (\texttt{Case3\_Transmission\_k0\_8}) for the GridNFMM and TFNO, where both the scattering medium $n(r)$ and incident plane wave $u^i(r)$ vary simultaneously.

\begin{figure}[ht!]
  \centering
  \setlength{\tabcolsep}{4pt}
  \begin{tabular}{@{} c c c @{}}
    \textbf{Input and Target} & \textbf{Prediction} & \textbf{Absolute Error} \\[0.3em]
    \includegraphics[width=0.315\linewidth]{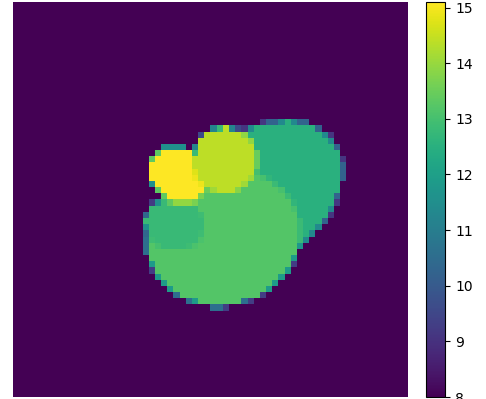} &
    \includegraphics[width=0.315\linewidth]{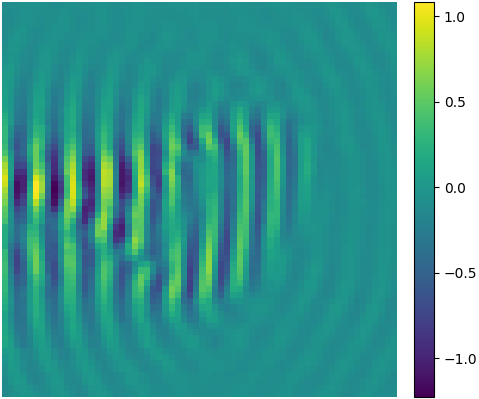} &
    \includegraphics[width=0.315\linewidth]{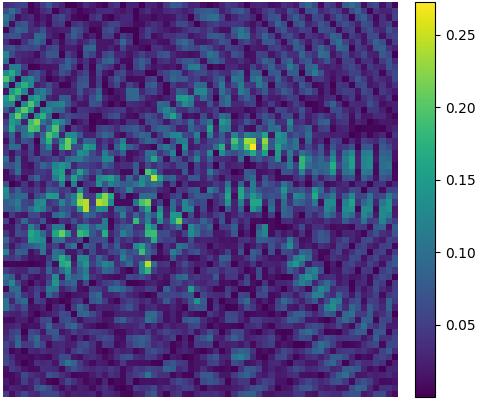} \\
    \footnotesize Input $f(r)$ & \footnotesize GridNFMM Prediction & \footnotesize GridNFMM Abs.\ Error \\[0.5em]
    \includegraphics[width=0.315\linewidth]{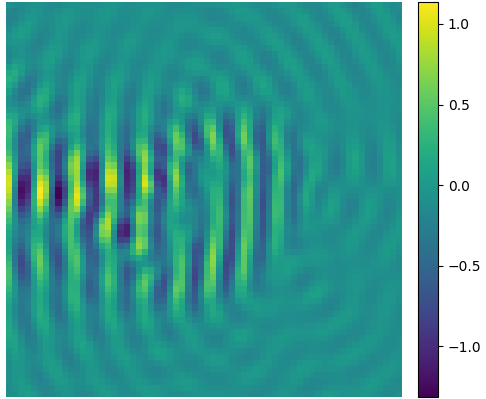} &
    \includegraphics[width=0.315\linewidth]{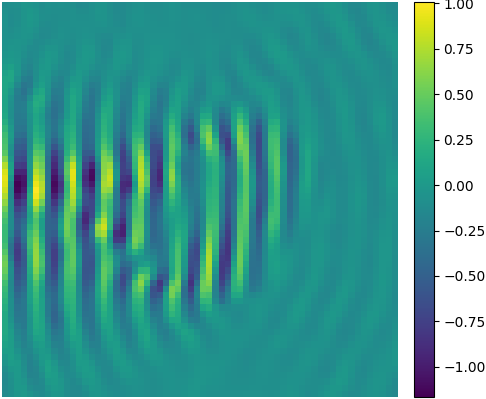} &
    \includegraphics[width=0.315\linewidth]{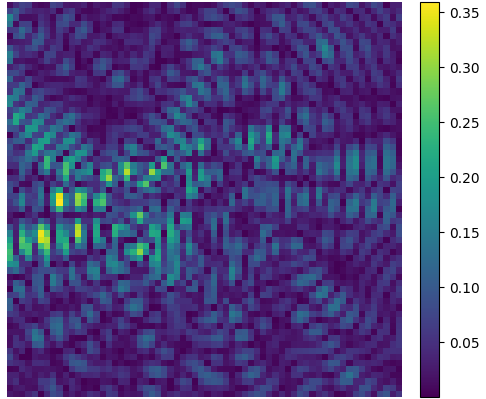} \\
    \footnotesize Target $u^s(r)$ & \footnotesize TFNO Prediction & \footnotesize TFNO Abs.\ Error
  \end{tabular}
  \caption{Comparison on \texttt{Case3\_Transmission\_k0\_8} with a plane-wave input: Shared input and target (left column), GridNFMM prediction and absolute error (top right), and TFNO prediction and absolute error (bottom right).}
  \label{fig:case3_transmission_k0_8_2x3}
\end{figure}

\FloatBarrier
% =============================================================================
\section{Proof of computational complexity}\label{apdx:complexity_proof}
% =============================================================================

Here we provide the proof of \cref{thm:gridnfmm-complexity} establishing the $\mathcal{O}(N)$ computational and memory complexity of the GridNFMM block (\cref{alg:gridnfmm}).

\begin{proof}[Proof of \cref{thm:gridnfmm-complexity}]
  We count floating-point operations level by level across the steps of \cref{alg:gridnfmm}, using the fact that every operator in the block performs $\mathcal{O}(1)$ matrix-vector or elementwise operations per box (or per interaction edge) at its level.

  First, since the quadtree is balanced with $M_\ell = 4^\ell$ boxes at level $\ell$ and $M_L = N$, the total number of boxes across all active tree levels forms a geometric series bounded linearly in $N$:
  \begin{equation}\label{eq:tree-size}
    \sum_{\ell=2}^{L} M_\ell
    \;=\; \sum_{\ell=2}^{L} 4^{\ell}
    \;\le\; \frac{4}{3}\,N
    \;=\; \mathcal{O}(N).
  \end{equation}
  At the leaf level, the encoding ($\ofs{}$), decoding ($\tfi{}$), gating, and GLU updates act on the $N$ leaves via dense channel projections of cost $\mathcal{O}(c^2)$ per leaf box, requiring $\mathcal{O}(N c^2)$ operations in total.
  Similarly, the upward ($\ofo{}$) and downward ($\ifi{}$) shift operators apply $c_T \times c_T$ triangular linear maps and RoPE rotations costing $\mathcal{O}(M_\ell c^2)$ at level $\ell$; summing over levels $\ell \in \{2, \dots, L\}$ and applying \cref{eq:tree-size} yields $\mathcal{O}(N c^2)$.

  For the far-field translation stage, the factorised $\ifo{}$ operator \cref{eqn:ifo_matrix_decomp} never materialises a dense $M_\ell \times M_\ell$ matrix.
  At level $\ell$, it applies the shared $c_T \times c_T$ matrices $\mathbf{W}_V, \mathbf{W}_O, \mathbf{W}_G$ once per box at cost $\mathcal{O}(M_\ell c^2)$, and gathers diagonal scalings $\mathbf{\Theta}_{\delta(\tau, \sigma)}$ along the precomputed interaction list $\mathcal{E}_\ell$.
  In two dimensions, the FMM well-separatedness criterion bounds the interaction list of each box by $|\mathcal{U}_\tau| \le 27$, so $|\mathcal{E}_\ell| \le 27 M_\ell = \mathcal{O}(M_\ell)$ and each edge costs $\mathcal{O}(c)$ for the diagonal multiplication and accumulation.
  Summing over all levels therefore requires $\sum_{\ell=2}^L \mathcal{O}(M_\ell c^2 + |\mathcal{E}_\ell|\, c) = \mathcal{O}(N c^2)$ operations.

  Finally, the leaf-level near-field CNN \cref{eqn:grid_leaf_cnn} gathers a fixed $3\times 3$ neighbourhood for each leaf box and applies a shared $c_L \to r$ compression ($r = \max(1, \lfloor c_L / 4 \rfloor)$), a $9r \to c_L$ spatial mixer, and a $c_L \to c_L$ output mixer, costing $\mathcal{O}(9\, c^2) = \mathcal{O}(c^2)$ per leaf, or $\mathcal{O}(N c^2)$ overall.
  Combining all four contributions yields
  \begin{equation*}
    \underbrace{\mathcal{O}(N c^2)}_{\text{Leaf Ops}}
    \;+\; \underbrace{\mathcal{O}(N c^2)}_{\text{Tree Shifts}}
    \;+\; \underbrace{\mathcal{O}(N c^2)}_{\text{Far-Field IFO}}
    \;+\; \underbrace{\mathcal{O}(N c^2)}_{\text{Near-Field CNN}}
    \;=\; \mathcal{O}\!\left(N c^2\right).
  \end{equation*}
  Treating the channel width $c$ as fixed independently of $N$, the block runs in $\mathcal{O}(N)$ time, with activation memory $\mathcal{O}(N c)$ plus $\mathcal{O}\!\left(\sum_{\ell=2}^L |\mathcal{E}_\ell|\right) = \mathcal{O}(N)$ for the cached sparse interaction indices.
\end{proof}

\begin{remark}
  Linear scaling relies on two structural assumptions: that the quadtree is balanced so that \cref{eq:tree-size} holds, and that the interaction masks are sparse with a bounded interaction list as guaranteed by the standard FMM admissibility criterion.
  If the interaction masks were dense, the far-field translation stage would degenerate to $\mathcal{O}(N^2)$.
\end{remark}

\bibliographystyle{siamplain}
\bibliography{bibliography}

\end{document}